\documentclass[10pt,twocolumn,letterpaper]{article}

\usepackage{cvpr}
\usepackage{times}
\usepackage{epsfig}
\usepackage{graphicx}
\usepackage{amsmath}
\usepackage{amssymb}
\usepackage{multirow}
\usepackage{subfigure}
\usepackage{setspace}
\usepackage{color}

\usepackage[breaklinks=true,bookmarks=false]{hyperref}

\cvprfinalcopy 

\def\cvprPaperID{4304} 

\ifcvprfinal\fi
\begin{document}

\title{Unsupervised Sparse Dirichlet-Net for Hyperspectral Image Super-Resolution}

\author{Ying Qu\textsuperscript{1}*,\qquad Hairong Qi\textsuperscript{1}, \qquad Chiman Kwan\textsuperscript{2}\\
\textsuperscript{1}The University of Tennessee, Knoxville, TN \qquad \textsuperscript{2} Applied Research LLC, Rockville, MD\\
{\tt\small yqu3@vols.utk.edu \qquad hqi@utk.edu \qquad chiman.kwan@arllc.net}}

\maketitle
\thispagestyle{empty}

\begin{abstract}
In many computer vision applications, obtaining images of high resolution in both the spatial and spectral domains are equally important. However, due to hardware limitations,
one can only expect to acquire images of high resolution in either the spatial or spectral domains. This paper focuses on hyperspectral image super-resolution (HSI-SR), where a hyperspectral image (HSI) with low spatial resolution (LR) but high spectral resolution is fused with a multispectral image (MSI) with high spatial resolution (HR) but low spectral resolution to obtain HR HSI. Existing deep learning-based solutions are all supervised that would need a large training set and the availability of HR HSI, which is unrealistic. Here, we make the first attempt to solving the HSI-SR problem using an unsupervised encoder-decoder architecture that carries the following uniquenesses. First, it is composed of two encoder-decoder networks, coupled through a shared decoder, in order to preserve the rich spectral information from the HSI network. Second, the network encourages the representations from both modalities to follow a sparse Dirichlet distribution which naturally incorporates the two physical constraints of HSI and MSI. Third, the angular difference between representations are minimized in order to reduce the spectral distortion. We refer to the proposed architecture as unsupervised Sparse Dirichlet-Net, or uSDN. Extensive experimental results demonstrate the superior performance of uSDN as compared to the state-of-the-art.
\end{abstract}

\section{Introduction}
\label{sec:intro}
\begin{figure}[t]
	\begin{center}
		{\includegraphics[width=0.8\linewidth]{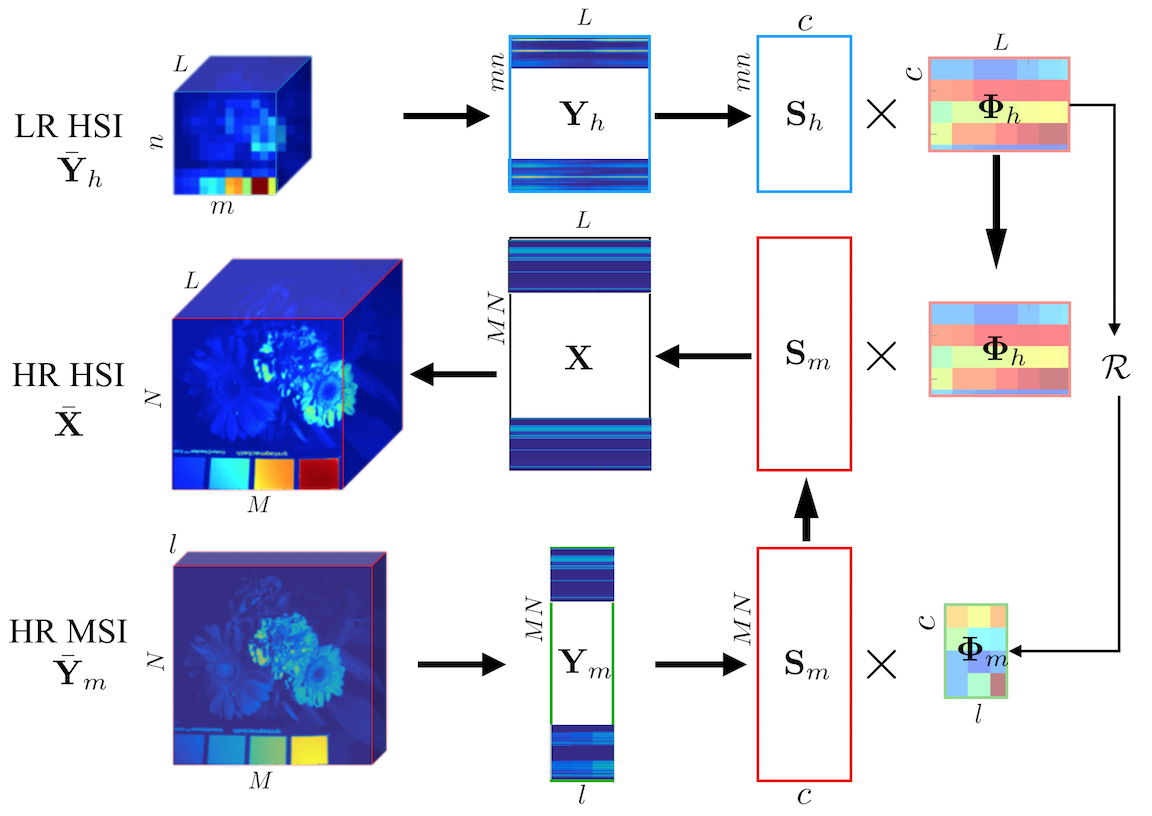}}
	\end{center}
	\vspace{-3mm}
	\caption{General procedure of HSI-SR.}
	\label{fig:procedure}
	\vspace{-3mm}
\end{figure}

Hyperspectral image (HSI) analysis has become a thriving and active research area in computer vision with a wide range of applications~\cite{chakrabarti2011statistics,bioucas2012hyperspectral}, including, for example, object recognition and classification~\cite{kwan2006novel, fauvel2013advances,zhang2016scene, maggiori2017recurrent}, tracking~\cite{van2010tracking,Fu_2016_CVPR, Uzkent_2016_CVPR_Workshops,Uzkent_2017_CVPR_Workshops}, environmental monitoring~\cite{spangler2010shallow,plaza2011foreword}, and change detection~\cite{kwon2005kernel,borengasser2007hyperspectral}. Compared to multispectral images (MSI with around 10 spectral bands) or conventional color images (RGB with 3 bands), HSI collects hundreds of contiguous bands which provide finer details of spectral signature of different materials. However, its spatial resolution becomes significantly lower than MSI or RGB due to hardware limitations~\cite{kawakami2011high,akhtar2015bayesian}. On the contrary, although MSI or RGB has high spatial resolution, their spectral resolution is relatively low. Very often, to yield better recognition and analysis results, images with both high spectral and spatial resolution are desired~\cite{vivone2015critical}. A natural way to generate such images is to fuse hyperspectral images with multispectral images or conventional color images. This procedure is referred to as \textit{hyperspectral image super-resolution (HSI-SR)}~\cite{akhtar2015bayesian,lanaras2015hyperspectral,dian2017hyperspectral} as shown in Fig.~\ref{fig:procedure}. 

The problem of HSI-SR originates from \textit{multispectral pan-sharpening (MSI-PAN)} in the remote sensing field, where the spatial resolution of MSI is further improved by a high-resolution panchromatic image (PAN). Note that, in general, “resolution” refers to the “spatial resolution.” Usually, MSI has much higher resolution than HSI, but PAN has even higher resolution than MSI. We use LR to denote low spatial resolution and HR for high spatial resolution. There are roughly two groups of MSI-PAN methods, namely, the component substitution (CS) ~\cite{thomas2008synthesis,chavez1991comparison,aiazzi2007improving} and the multi-resolution analysis (MRA) based approaches~\cite{aiazzi2006mtf}. Although MSI-PAN has been well developed through decades of innovations~\cite{thomas2008synthesis,loncan2015hyperspectral,zhou2016hyperspectral}, they cannot be readily adopted to solve the HSI-SR problems. On one hand, the amount of spectral information to be preserved for HSI-SR is much higher than that of MSI-PAN, thus it is easier to introduce spectral distortion, \ie, the output image does not preserve the accurate spectral information~\cite{loncan2015hyperspectral,yokoya2017hyperspectral, akhtar2015bayesian,dian2017hyperspectral}. On the other hand, HSI possesses much lower resolution than that of MSI, making it more challenging to improve the spatial resolution. 

There have been few methods specifically designed for HSI-SR, including mainly Bayesian based and matrix factorization based approaches~\cite{loncan2015hyperspectral,yokoya2017hyperspectral,dong2016hyperspectral}. The unique framework of Bayesian offers a convenient way to regularize the solution space of HR HSI by employing a proper prior distribution such as Gaussian ~\cite{wei2015hyperspectral}. Simoes \etal proposed HySure~\cite{simoes2015convex}, which applied a total variation regularization to smooth the image. Akhtar  \etal~\cite{akhtar2015bayesian} introduced a non-parametric Bayesian strategy to extract spectral dictionary and spatial coefficients from LR HSI and HR MSI, respectively. Matrix factorization based approaches have been actively studied recently~\cite{kawakami2011high,yokoya2012coupled,dong2016hyperspectral,lanaras2015hyperspectral,veganzones2016hyperspectral}. Yokoya \etal~\cite{yokoya2012coupled} decomposed both the LR HSI and HR MSI alternatively to achieve the optimal non-negative bases and coefficients that used to generate HR HSI. Lanaras \etal~\cite{lanaras2015hyperspectral} further improved the fusion results by introducing a sparse constraint. However, most existing HSI-SR approaches generally assume that the downsampling function between the spatial coefficients of HR HSI and LR HSI are known beforehand. This assumption is not always true due to the distortions caused by both the sensors and complex environmental conditions~\cite{akhtar2015bayesian}. 

HSI-SR is also closely related to the natural image super-resolution (SR) problem, which has been extensively studied and achieved excellent performance through the state-of-the-art \textit{deep learning}~\cite{dong2016image, Lu_2015_CVPR, Shi_2016_CVPR, Kim_2016_CVPR1,Kim_2016_CVPR2, ledig2016photo, lai2017deep, he2016deep}. The main principle of SR is to learn a mapping function between LR images and HR images in a supervised fashion. Natural image SR methods usually work on up to $4\times$ upscaling. There have been three attempts to address the MSI-PAN problem with deep learning where the mapping function is learned using different frameworks including tied-weights denoising/ autoencoder~\cite{huang2015new}, SRCNN~\cite{masi2016pansharpening}, and deep residual network~\cite{he2016deep,wei2017boosting}. These deep learning based methods, including natural image SR and MSI-PAN are all supervised, making their adoption on HSI-SR a challenge due to two reasons. First, they are designed to find an end-to-end mapping function between the LR images and HR images under the assumption that the mapping function is the same for different images. However, the mapping function may not be the same for images acquired with different sensors. Even for the data collected from the same sensor, the mapping function for different spectral bands may not be the same. Thus the assumption may cause severe spectral distortion. Second, training a mapping function is a supervised solution which requires a large dataset, the down-sampling function, and the availability of the HR HSI, that are not realistic for HSI.

\begin{figure*}[t]
	\begin{center}
		\begin{minipage}{0.35\textwidth}
			\includegraphics[width=1\linewidth]{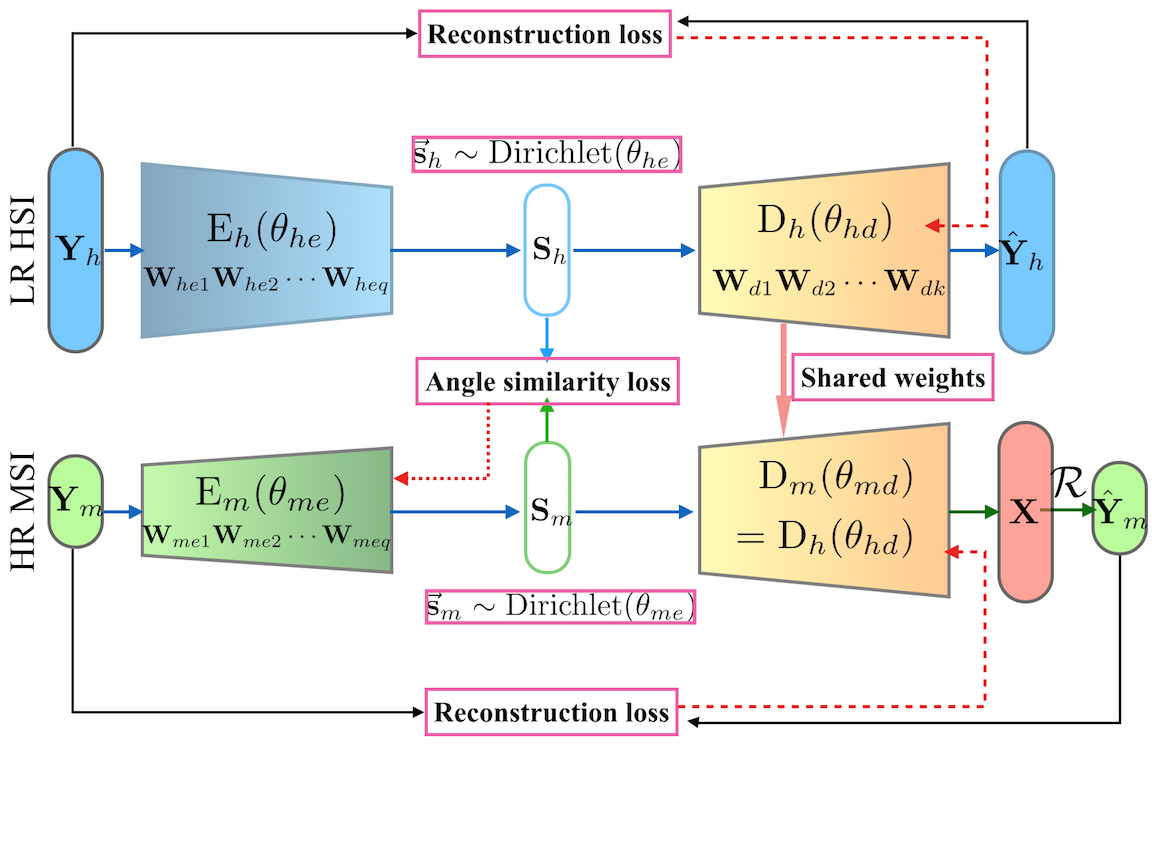}
			\caption{Simplified architecture of uSDN.}
			\label{fig:simple}
		\end{minipage}\hspace{12mm}
		\begin{minipage}{0.55\textwidth}
			{\includegraphics[width=1\linewidth]{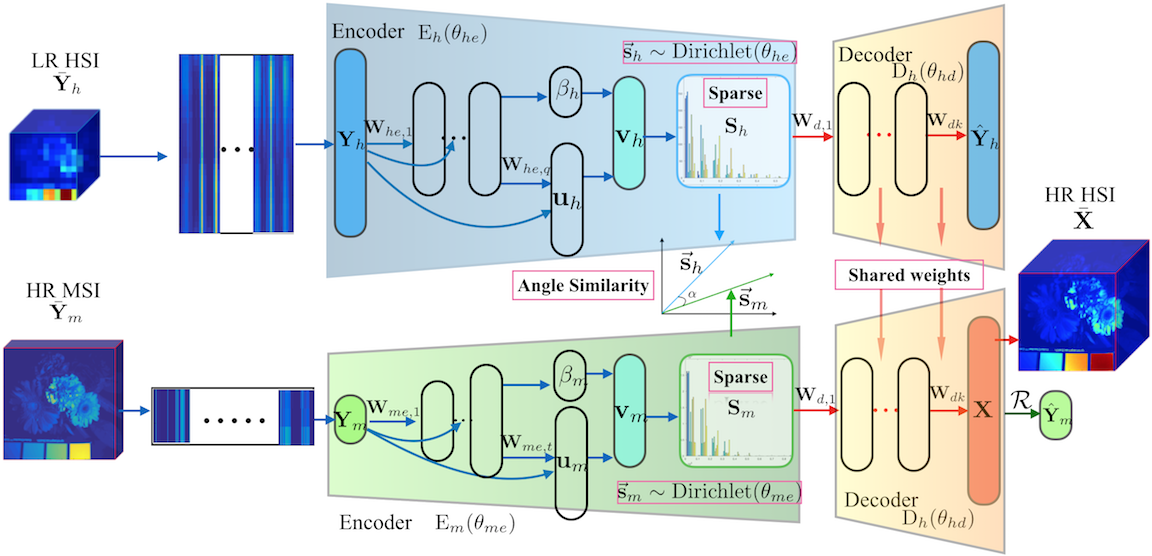}}
			\caption{Details of the encoder nets.}
			\label{fig:flow}
		\end{minipage}\hfill
	\end{center}
\vspace{-5mm}
\end{figure*}

In this paper, we propose an \textit{unsupervised} network structure to address the challenges of HSI-SR. To the best of our knowledge, this is the first effort to solving the HSI-SR problem with deep learning in an unsupervised fashion. The novelty of this work is three-fold. First, the network extracts both the spectral and spatial information from LR HSI and HR MSI with two deep learning networks which share the same decoder weights, as illustrated in  Fig.~\ref{fig:simple}. Second, in order to incorporate the two physical constraints of HSI and MSI data representation, i.e., sum-to-one and sparsity, the network encourages the representations from both modalities to follow a Dirichlet distribution which naturally incorporates the sum-to-one property. Since each pixel of the image only consists of a few spectral bases, the sparsity of the representations is guaranteed by minimizing their entropy function. Third, to address the challenge of spectral distortion, instead of adopting the down-sampling function (as an estimated mapping function) to relate the representations of the two modalities, we minimize the angular difference of these representations such that they have similar patterns. In this way, the spectral distortion is largely reduced. The proposed method is referred to as uSDN. 

\section{Problem Formulation}
\label{sec:formulate}
Given the LR HSI, $\bar{\mathbf{Y}}_h \in \mathbb{R}^{m\times n\times L}$, where $m$, $n$ and $L$ denote the width, height and number of spectral bands of the HSI, respectively,
and the corresponding HR MSI, $\bar{\mathbf{Y}}_m \in \mathbb{R}^{M \times N \times l}$, where $M$, $N$ and $l$ denote the width, height and number of spectral bands of the MSI, respectively, the goal is to estimate the HR HSI,  $\bar{\mathbf{X}} \in \mathbb{R}^{M\times N \times L}$,  with both high spatial and spectral resolution. In general, MSI has much higher spatial resolution than HSI, \ie, $M\gg m$, $N \gg n$, and HSI has much higher spectral resolution than MSI, \ie, $L \gg l$. To facilitate the subsequent processing, we unfold the 3D images into 2D matrices, \ie, each row of the 2D matrix denotes the spectral reflectance of a given pixel. The unfolded matrices are written as $\mathbf{Y}_h \in \mathbb{R}^{mn\times L}$, $\mathbf{Y}_m \in \mathbb{R}^{MN \times l}$ and $\mathbf{X}\in \mathbb{R}^{MN \times L}$. This is illustrated in Fig.~\ref{fig:procedure}.

Assuming that each row of $\mathbf{Y}_h$ is a linear combination of $c$ basis vectors (or spectral signatures), as expressed in Eq. \eqref{equ:blrhsi}, where $\mathbf{\Phi}_h\in\mathbb{R}^{c \times L}$ and each row of which  denotes the spectral basis that preserves the spectral information and $\mathbf{S}_h\in\mathbb{R}^{mn\times c}$ is the corresponding proportional coefficients (referred to as \textit{representations} in deep learning). Since the coefficients indicate how the spectral bases are mixed at specific spatial locations, they preserve the spatial structure of HSI.

Similarly, $\mathbf{Y}_m$ can be expressed as Eq. \eqref{equ:bmsi}, where $\mathbf{\Phi}_m\in\mathbb{R}^{c \times l}$ and each row of which indicates the spectral basis of MSI. $\mathcal{R}\in\mathbb{R}^{L \times l}$ is the transformation matrix given as a prior from the sensor \cite{kawakami2011high, yokoya2012coupled, wei2015hyperspectral,loncan2015hyperspectral,simoes2015convex,vivone2015critical,lanaras2015hyperspectral,dian2017hyperspectral}, which describes the relationship between HSI and MSI bases. With $\mathbf{\Phi}_h\in\mathbb{R}^{c \times L}$ carrying the high spectral information and $\mathbf{S}_m \in\mathbb{R}^{MN\times c}$ carrying the high spatial information, the desired HR HSI, $\mathbf{X}$, is generated by Eq. \eqref{equ:bhrhsi}. See Fig.~\ref{fig:procedure}.

\begin{align}
\begin{split}\label{equ:blrhsi}
&\mathbf{Y}_h = \mathbf{S}_h\mathbf{\Phi}_h,
\end{split}\\
\begin{split}\label{equ:bmsi}
&\mathbf{Y}_m = \mathbf{S}_m\mathbf{\Phi}_m ,\quad \mathbf{\Phi}_m =\mathbf{\Phi}_h\mathcal{R}
\end{split}\\
\begin{split}\label{equ:bhrhsi}
&\mathbf{X} = \mathbf{S}_m\mathbf{\Phi}_h.
\end{split}\end{align}

The problem of HSI-SR can be described mathematically as $P(\mathbf{X}\vert\mathbf{Y}_h,\mathbf{Y}_m)$. Since the ground truth $\mathbf{X}$ is not available, the problem should be solved in an unsupervised fashion. The key to addressing this problem is to take advantage of the shared information, \ie, $\mathbf{\Phi}_h\in\mathbb{R}^{c \times L}$, to extract desired high spectral bases $\mathbf{\Phi}_h$ and spatial representations $\mathbf{S}_m$ from two different modalities. 

In addition, three unique requirements of HSI-SR need to be given special consideration. First, in representing HSI or MSI as a linear combination of spectral signatures, the representation vectors should be non-negative and sum-to-one. That is, $\sum_{j=1}^{c}s_{ij}=1$, where $\mathbf{s}_i$ is the row vector of either $\mathbf{S}_h$ or $\mathbf{S}_m$ \cite{kawakami2011high,yokoya2012coupled,dong2016hyperspectral,lanaras2015hyperspectral,veganzones2016hyperspectral}. Second,  due to the fact that each pixel of image only consists of a few spectral bases, the representations should be sparse. Third, spectral distortion should be largely reduced in the process in order to preserve the spectral information of HR HSI while gaining spatial resolution.

\section{Proposed Approach}
\label{sec:proposed}
We propose an unsupervised architecture as shown in Fig.~\ref{fig:simple}. We highlight the three structural uniquenesses here. First, the architecture consists of two deep networks, for the representation learning of the LR HSI and HR MSI, respectively. These two networks share the same decoder weights, enabling the extraction of both spectral and spatial information from multi-modalities in an unsupervised fashion. Second, in order to satisfy the sum-to-one constraint of the representations, both $\mathbf{S}_h$ and $\mathbf{S}_m$ are encouraged to follow a Dirichlet distribution where the sum-to-one property is naturally incorporated in the network with a further sparsity constraint. Third, to address the challenge of spectral distortion, the representations of two modalities are encouraged to have similar patterns by minimizing their angular difference. 

\subsection{Network Architecture}
\label{sec:arch}
As shown in Fig.~\ref{fig:simple}, the network reconstructs both the LR HSI $\mathbf{Y}_h$ and HR MSI $\mathbf{Y}_m$ in a coupled fashion. Taking the LR HSI network (the top network) as an example. The network consists of an encoder $\text{E}_h(\theta_{he})$, which maps the input data to low-dimensional representations (latent variables on the Bottleneck hidden layer), \ie, $p_{\theta_{he}}(\mathbf{S}_h\vert \mathbf{Y}_h)$, and a decoder  $\text{D}_h(\theta_{hd})$ which reconstructs the data from the representations, \ie, $p_{\theta_{hd}}(\hat{\mathbf{Y}}_h \vert \mathbf{S}_h)$. Both the encoder and decoder are constructed with multiple fully-connected layers. Note that the bottleneck hidden layer $\mathbf{S}_h$ behaves as the representation layer that reflect the spatial information and the weights $\theta_{hd}$ of the decoder $\text{D}_h(\theta_{hd})$ serve as $\mathbf{\Phi}_h$ in Eq.~\eqref{equ:blrhsi}, respectively. This correspondence is further elaborated below. 

The HSI is reconstructed by $\hat{\mathbf{Y}}_h =  f_k(\mathbf{W}_{dk}f_{k-1}(...(f_1(\mathbf{S}_h\mathbf{W}_{d1}+b_1)...)+b_{k-1})+b_k)$, where $\mathbf{W}_{dk}$ denotes the weights in the $k$th layer. To extract the spectral basis from LR HSI, the latent variables of the representation layer
$\mathbf{S}_h$ act as the proportional coefficients, where $\mathbf{S}_h$ follows a Dirichlet distribution with the sum-to-one property naturally incorporated. Suppose the activation function is an identity function and there is no bias in the decoder, we have $\theta_{hd} = \mathbf{W}_1\mathbf{W}_2...\mathbf{W}_k$. That is, the weights $\theta_{hd}$ of the decoder correspond to the spectral basis $\mathbf{\Phi}_h$ in Eq.~\eqref{equ:blrhsi} and $\mathbf{\Phi}_h = \theta_{hd} $ . In this way,  $\mathbf{\Phi}_h$  preserves the spectral information of LR HSI, and the latent variables $\mathbf{S}_h$ preserves the spatial information effectively.

Equivalently, the bottom network reconstructs the HR MSI in a similar way with encoder $\text{E}_m(\theta_{me})$ and decoder $\text{D}_m(\theta_{md})$. However, since $l\le c \le L$, \ie, the number of latent variables, $L$, is much larger than the number of input nodes, $l$, the MSI network is very unstable and hard to train. On the other hand, the spectral basis of HR MSI can be transformed from those of LR HSI which possesses more spectral information, the decoder of the MSI is designed to share the weights with that of HSI in terms of $\theta_{md} =\mathbf{\Phi}_m= \theta_{hd}\mathcal{R} = \mathbf{\Phi}_h\mathcal{R}$. Then the reconstructed HR MSI can be obtained by $\hat{\mathbf{Y}}_m = \mathbf{S}_m\mathbf{\Phi}_h\mathcal{R}$. In this way, only the encoder $\text{E}_m(\theta_{me})$ of the MSI is updated during the optimization, where the HR spatial information $\mathbf{S}_m$ is extracted from MSI.  Eventually, the desired HR HSI is generated directly by $\mathbf{X} = \mathbf{S}_m\mathbf{\Phi}_h$. Note that the dashed lines in the image show the path of backpropagation which will be elaborated in Sec.~\ref{sec:label}.

\subsection{Sparse Dirichlet-Net with Dense Connectivity}
\label{sec:diri}
To extract stable spectral information, we need to enforce the proportional coefficients $\mathbf{S} = (\mathbf{s}_1, \mathbf{s}_2, \cdots, \mathbf{s}_i, \cdots, \mathbf{s}_p)^T$ of each pixel to sum-to-one \cite{yokoya2012coupled,wycoff2013non,lanaras2015hyperspectral,lanaras2015hyperspectral}, \ie, $\sum_{j=1}^{c}s_{ij}=1$. Without loss of generality, $\mathbf{S}$ represents either $\mathbf{S}_h$ with $p=mn$ or $\mathbf{S}_m$ with $p=MN$. In addition, due to the fact that only a few spectral bases actually contribute in the linear combination of the spectral reflectance of each pixel, the coefficients should also be sparse. In the proposed architecture, the latent variables (or representations) of the hidden layer $\mathbf{S}_h$ or $\mathbf{S}_m$ correspond to the proportional coefficients  in Eqs.~\eqref{equ:blrhsi} and \eqref{equ:bmsi}. To naturally incorporate the sum-to-one property, the representations are encouraged to follow a Dirichlet distribution which is accomplished with stick-breaking process as illustrated in Fig. 3.  Furthermore, entropy function is adopted to reinforce the sparsity of the representations.

The stick-breaking process was first proposed by Sethuranman \cite{sethuraman1994constructive} back in 1994. It is used to generate random vectors $\mathbf{s}$ with Dirichlet distribution. The process can be illustrated as breaking a unit-length stick into $c$ pieces, the length of which follows a Dirichlet distribution.  Assuming that the generated vector is denoted as $\mathbf{s} = (s_1,\cdots,s_j,\cdots,s_c)$, we have $0\leq s_j\leq1$, and the variables in the vector are sum to one, \ie, $\sum_{j=1}^{c}{s_j}=1$.  Mathematically \cite{sethuraman1994constructive}, a single variable ${s_j}$ is defined as
\begin{equation}
s_j =\left\{
\begin{array}{ll}
v_1 \quad & \text{for} \quad j = 1\\
v_j\prod_{o<j}(1-v_o) \quad &\text{for} \quad j>1 , 
\end{array}\right.
\label{equ:stick}
\end{equation}
where $v_o$ is drawn from a Beta distribution, \ie, $v_o\sim \text{Beta}(u, \alpha,\beta)$. 
Nalisnick and Smyth successfully coupled the expressiveness of generative networks with Bayesian nonparametric model through stick-breaking process \cite{nalisnick2016deep}. The network uses a Kumaraswamy distribution \cite{kumaraswamy1980generalized} as an approximate posterior which takes in the samples from a randomly generated uniform distribution during the training procedure. 

Different from the generative network, we aim to find shared representations that better reconstruct the data. Therefore, the weights of the network should be changed according to the input data instead of randomly generated distribution. It has been proved that when $v_o\sim \text{Beta}(u, 1,\beta)$, $\mathbf{s}$ follows a Dirichlet distribution. Since it is difficult to draw samples directly from Beta distribution, we draw samples from the inverse transform of Kumaraswamy distribution, as shown in Eq.~\eqref{equ:kuma}, which is equivalent to Beta distribution when $\alpha = 1$ or $\beta = 1$, 
\begin{equation}
\text{kuma}(u,\alpha,\beta) = \alpha\beta u^{\alpha-1}(1-u^{\alpha})^{\beta-1}
\label{equ:kuma}
\end{equation}
where $\alpha>0$, $\beta>0$ and $u \in (0,1)$. 
The benefit of Kumaraswamy distribution is that it has a closed-form CDF, where the inverse transform is  defined as 
\begin{equation}
v_o\sim (1-(1-u^\frac{1}{\beta})^\frac{1}{\alpha}).
\label{equ:draw}
\end{equation}

Let $\alpha = 1$, parameters $u$ and $\beta$ are learned through the network as illustrated in Fig.~\ref{fig:flow}. Because $\beta>0$, a softplus is adopted as the activation function \cite{dugas2001incorporating} at the ${\beta}$ layer. Similarly, a sigmoid \cite{han1995influence} is used to map ${u}$ into $(0,1)$ range at the $\mathbf{u}$ layer. To avoid gradient vanishing and increase the representation power of the proposed method, the encoder of the network is densely connected, \ie, each layer is fully connected with all its subsequent layers \cite{huang2016densely}. 

To further increase the variability of $u$ and $\beta$ (theoretically, we want the learned $u$ and $\beta$ to be any number within their range), instead of concatenating all the preceding layers, the input of the $k$th layer is the summation of all the preceding layers $x_0,~x_1,~x_{k-1}$ with their own weights, \ie, $\mathbf{W}_0x_0 + \mathbf{W}_1x_1 + ...+\mathbf{W}_{k-1}x_{k-1}$. In this way, fewer number of layers is required to learn the optimal representations. 

Although the stick-breaking structure encourages the representations to follow a Dirichlet distribution, it does not guarantee the sparsity of the representations. In addition, the widely used $l_1$ regularization or Kullback-Leibler divergence \cite{Goodfellow-et-al-2016} will not encourage the representation layer to be sparse either, because they guarantee the sparsity by reducing the mean of active value, \ie, mean of the representation layer. However, due to the stick-breaking structure, the mean of $\mathbf{S}_h$ or $\mathbf{S}_m$ is almost one. Therefore, we introduce a generalized Shannon entropy function~\cite{huang2017sparse} to reinforce the sparsity of the representation layer which works effectively even with the sum-to-one constraint. 

\begin{figure}[t]
	\begin{center}
		\includegraphics[width=0.35\linewidth]{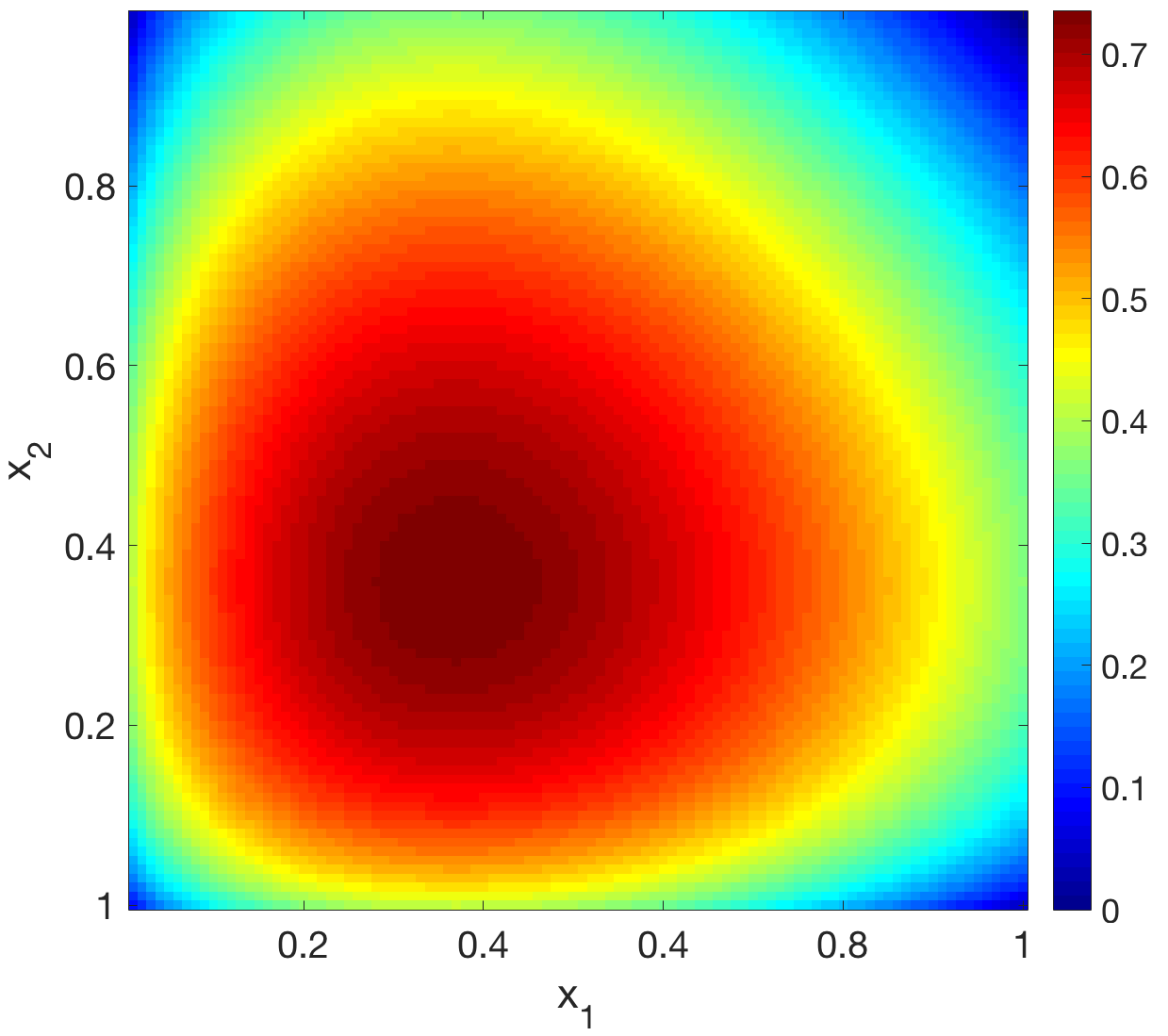}\hspace{4mm}\hspace{5mm}
		\includegraphics[width=0.35\linewidth]{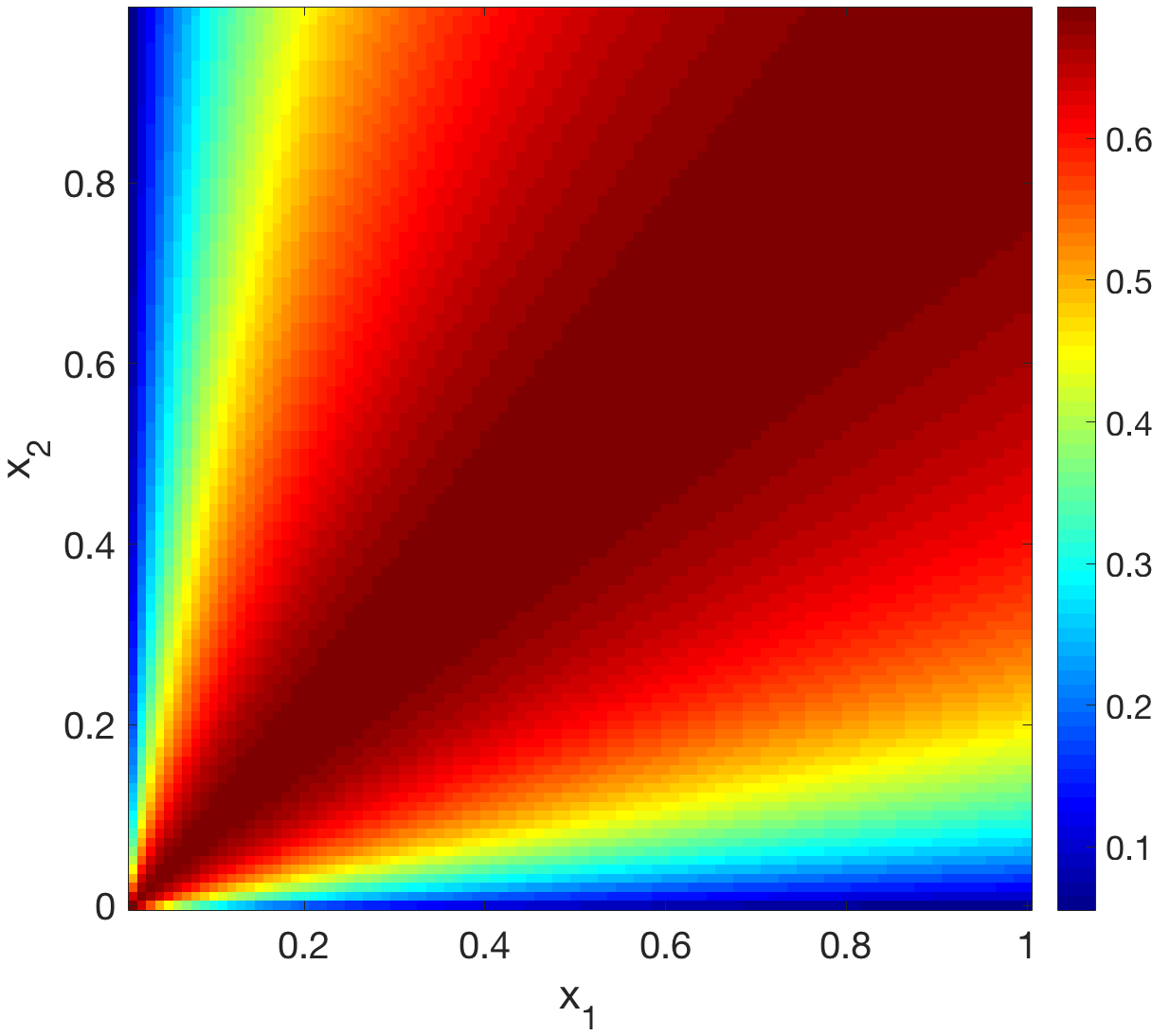}\hfill
	\end{center}
	\vspace{-4mm}
	\caption{Shannon entropy (L) and Shannon entropy function (R).}
	\label{fig:entropy}
	\vspace{-4mm}
\end{figure}

The entropy function was first proposed in compressive sensing field to solve the signal recovery problem. It is defined as
\begin{equation}
\mathcal{H}_{p}(\mathbf{s}) = -\sum_{j=1}^{N}\frac{\vert s_j \vert^p}{\Vert \mathbf{s} \Vert_p^p}
\log\frac{\vert s_j \vert^p}{\Vert {\mathbf{s}} \Vert_p^p}. 
\label{equ:entropyfun}
\end{equation}

Compared to the more popular Shannon entropy, the entropy function Eq.~\eqref{equ:entropyfun} decreases monotonically when the data become sparse. To illustrate the effect, we show the phenomena with 2D variables in Fig.~\ref{fig:entropy}. Shannon entropy is small when both $x_1$ and $x_2$ are small or large. But for Shannon entropy function, 
the local minimum only occurs at the boundaries of the quadrants. This nice property guarantees the sparsity of arbitrary data even the data are with the sum-to-one constraint. Due to the stick-breaking structure, the latent variables at the representation layer are positive. We choose $p=1$ which is more efficient and will encourage the variables to be sparse. 

\subsection{Angle Similarity}
\label{sec:angle}
Extracting spatial information from HR MSI is quite challenging and easy to introduce spectral distortion in the subsequent HR HSI results. The main cause to this problem is that the number of the representations $c$ (number of nodes in the representation layer) is much larger than the dimension of the MSI, \ie, $c \gg l$. Previous researchers assume the down-sampling function is available \textit{a-priori} to build a relationship between the representations of HSI and MSI. However, the down-sampling function is usually unknown for real applications. 

Therefore, instead of taking the down-sampling function as a prior, we encourage the representations $\mathbf{S}_h$ and $\mathbf{S}_m $ of the two networks following a similar pattern to prevent spectral distortion. And such similarity is measured by the angular difference between the two representations. Spectral angle mapper (SAM) is employed to measure this angular difference. SAM is a spectral evaluation method in remote sensing \cite{loncan2015hyperspectral, yokoya2017hyperspectral, endnet2017}, which measures the angular difference between the estimated image and the ground truth image. The lower the SAM score, the smaller the spectral angle difference, and the more similar the two representations. 

Since the HSI and MSI networks share the same decoder weights, the representations should have similar angle in order to generate high quality image with less spectral distortion. Besides encouraging the representation layer to follow a sparse Dirichlet distribution, we further reduce the angular difference of the representations of HSI and MSI during the optimization procedure.

In the network, representations ${\mathbf{{S}}}_h \in\mathbb{R}^{mn \times c}$ and ${\mathbf{{S}}}_m\in\mathbb{R}^{MN \times c}$, from two different modalities have different dimensions. To minimize the angular difference, we increase the size of the low-dimensional ${\mathbf{{S}}}_h$ by duplicating its values at each pixel to its nearest neighborhood. Then the duplicated representations $\tilde{{\mathbf{{S}}}_h}\in\mathbb{R}^{MN\times c}$ have the same dimension as ${\mathbf{{S}}}_m$. With vectors of equal size, the angular difference is defined as 
\begin{equation}
\mathcal{A}(\tilde{{\mathbf{{S}}}_h}, {\mathbf{{S}}}_m )= \frac{1}{MN} \sum_{i=1}^{MN}\arccos(\frac{\tilde{\mathbf{s}}_h^{~i}\cdot \mathbf{s}_m^{~i}}{\Vert\tilde{\mathbf{s}}_h^{~i}\Vert_2\Vert\mathbf{s}_m^{~i} \Vert_2})
\label{equ:angle}
\end{equation}

To map the range of the angle within $(0,1)$, Eq.~\eqref{equ:angle} is divided by the circular constant $\pi$. 
\begin{equation}
\mathcal{J}(\tilde{\mathbf{S}_h}, \mathbf{S}_m ) = \frac{\mathcal{A}(\tilde{{\mathbf{{S}}}_h}, {\mathbf{{S}}}_m )}{\pi}
\end{equation}

\subsection{Optimization and Implementation Details}
\label{sec:label}
To prevent over-fitting, we applied an $l_2$ norm on the decoder weights. The objective functions of the proposed network architecture can then be expressed as:
\begin{align}
\begin{split}\label{equ:objhsi}
&\mathcal{L}(\theta_{he}, \theta_{hd}) = \frac{1}{2}\Vert\mathbf{Y}_h(\theta_{he}, \theta_{hd})-\hat{\mathbf{Y}}_h(\theta_{he}, \theta_{hd})\Vert_F^2\\
&+\lambda \mathcal{H}_1(\mathbf{S}_h(\theta_{he})) + \mu \Vert \theta_{hd}\Vert_F^2,
\end{split}\\
\begin{split}\label{equ:objmsi}
&\mathcal{L}(\theta_{me}) = \frac{1}{2}\Vert\mathbf{Y}_m(\theta_{me}, \theta_{hd})-\hat{\mathbf{Y}}_m(\theta_{me}, \theta_{hd})\Vert_F^2\\
& +\lambda \mathcal{H}_1(\mathbf{S}_m(\theta_{me})),
\end{split}\\
\begin{split}\label{equ:objangle}
&\mathcal{L}(\theta_{me}) = \mathcal{J}(\tilde{\mathbf{S}}_h(\theta_{he}), \mathbf{S}_m(\theta_{me})),
\end{split}
\end{align}
where $\lambda$ and $\mu$ are parameters that balance the trade-off between the reconstruction error and the sparsity and weights loss, respectively. 

The proposed architecture consists of two sparse Dirichlet-Nets which extract the spectral information $\Phi_h$ from HSI and spatial information $\mathbf{S}_m$ from MSI. The network is optimized with back-propagation following the procedure described below, also illustrated in Fig.~\ref{fig:simple} with the dashed line. 

Step 1: Since the decoder weights $\theta_{hd}$ of the HSI network preserves the spectral information $\Phi_h$, we first update the HSI network, given the objective function in Eq.~\eqref{equ:objhsi}, to find the optimal $\theta_{hd}$. To prevent over-fitting, an $l_2$ norm is applied on the decoder of the HSI network. 

Step 2: The estimated decoder weights $\theta_{hd}$ are fixed and shared with the decoder of the MSI network. Update the encoder weights $\theta_{me}$ of the MSI network given the objective function in Eq.~\eqref{equ:objmsi}. 

Step 3: To reduce spectral distortion, every 10 iterations, we minimize the angular difference between the representations of two modalities given the objective function in Eq.~\eqref{equ:objangle}. Since we already have $\theta_{he}$ from the first step, only the encoder $\theta_{me}$ of the MSI network is updated during the optimization. 

For all the experiments, both the input and output of the HSI network have 31 nodes, representing the number of spectral bands in the data. The numbers of densely-connected layers and nodes of the encoder are shown in Table~\ref{tab:layers}. There are 3 layers in the HSI network and each layer contains 10 nodes. The MSI network has 5 layers with the number of nodes increases from $4$ to $10$. The $\mathbf{v}_h$/$\mathbf{v}_m$  are drawn with Eq.~\eqref{equ:draw} given $\mathbf{u}_h$/$\mathbf{u}_m$ and $\beta_h$/$\beta_m$, which are learned by back-propagation. Both $\beta_h$ and $\beta_m$ have only one node, denoting the distribution parameter of each pixel. The representation layers, $\mathbf{S}_h$ and  $\mathbf{S}_m$  with 10 nodes are constructed with $\mathbf{v}_h$ and $\mathbf{v}_m$, respectively, according to Eq.~\eqref{equ:stick}. The network shares the decoder with 2 layers and each layer has 10 nodes. 
Since different images have different spectral bases and representations, the network is trained on each pair of LR HSI and HR MSI to reconstruct each image accurately. 
\begin{table}[htb]
	\caption{The number of layers and nodes in the network.}
	\label{tab:layers}
	\begin{center}
		\begin{tabular}{c|cccc}
			\hline
			\multirow{2}{*}{Dirichlet-Net}&\multicolumn{4}{|c}{Encoder} \\
			\cline{2-5}
			&$\#$layers and $\#$nodes&$\mathbf{u}$&${\beta}$&$\mathbf{v}$\\
			\hline
			HSI &	3 / [10,10,10]& 10&1&10\\		
			MSI &	5 / [4,5,7,910]& 10&1&10\\		
			\hline
		\end{tabular}
	\end{center}
\vspace*{-4mm}
\end{table}
\section{Experiments and Results}
\label{sec:experiment}
\noindent
\subsection{Datesets and Experimental Setup}
The proposed uSDN has been thoroughly evaluated with two widely used benchmark datasets, CAVE \cite{yasuma2010generalized} and Harvard \cite{chakrabarti2011statistics}. The CAVE dataset consists of 32 HR HSI images and each of which has a dimension of $512\times 512$ with 31 spectral bands. These spectral images are taken within the wavelength range 400 $\sim$ 700nm with an interval of 10 nm. The Harvard dataset includes 50 HR HSI images with both indoor and outdoor scenes. The dimension of the images in this dataset is $1392\times1040$, with 31 bands taken at an interval of 10nm within the wavelength range of 420 $\sim$ 720nm. Note that for this dataset, the top left corner of size $1024\times 1024\times 31$ is cropped as the HR HSI.

For the two benchmark datasets, the LR HSI $\mathbf{Y}_h$ is obtained by averaging the HR HSI over $32\times 32$ disjoint blocks. The HR MSI images with 3 bands are generated by multiplying the HR HSI with the given spectral response matrix $\mathcal{R}$ of Nikon D700. All the images are normalized between 0 and 1. Note that the CAVE dataset is in general considered a more challenging set than Harvard since images in Harvard usually contain more smooth reflections; and since the images have higher spatial resolution, pixels within close vicinity usually have similar spectral reflectance. Hence, even the images are down-sampled by the $32\times 32$ kernel, most spectral information is still preserved in the LR HSI.

The results of the proposed method on individual images are compared with seven state-of-the-art methods, \ie, CS based \cite{aiazzi2007improving}, MRA based \cite{aiazzi2006mtf}, CNMF \cite{yokoya2012coupled}, Bayesian Sparse (BS) \cite{wei2015hyperspectral}, HySure \cite{simoes2015convex}, Lanaras's 15 (CSU)  \cite{lanaras2015hyperspectral}, and Akhtar's 15 (BSR) \cite{akhtar2015bayesian}  that belong to different categories of approaches described in Sec. 1. These methods also reported the best performance \cite{loncan2015hyperspectral,akhtar2015bayesian,lanaras2015hyperspectral}, with the original code made available by the authors. 
We also directly list results \cite{akhtar2016hierarchical} from Akhtar's 16 (HBPG) since the code is not available. 
The average results on the complete dataset is also reported to evaluate the robustness of the proposed method. 
For quantitative comparison, the root mean squared error (RMSE) and spectral angle mapper (SAM) are applied to evaluate the reconstruction error and the amount of spectral distortion, respectively.

\subsection{Experimental Results} 
Tables \ref{tab:rmse} and \ref{tab:sam} show the experimental results of 7 groups of images from the CAVE and Harvard datasets, which are commonly benchmarked by existing literature \cite{kawakami2011high,akhtar2015bayesian,akhtar2016hierarchical}. We observe that traditional CS-based and MRA-based methods suffer from spectral distortion, thus could not achieve competitive performance. The Bayesian based approach, BS \cite{wei2015hyperspectral}, fails due to the fact that it assumes the representation $\mathbf{S}_m$ follows a Gaussian distribution, which is not always true. However, the Bayesian non-parametric based method BSR \cite{akhtar2015bayesian} outperforms BS because it estimates the spectra through non-parametric learning. The matrix-based approaches,CNMF \cite{yokoya2012coupled} and CSU \cite{lanaras2015hyperspectral}, are not as competitive on the CAVE dataset due to their predefined down-sampling function, although they perform much better on the Harvard dataset. We also observe that some methods like Hysure can achieve better RMSE, but worse SAM scores, that is because they cannot preserve the spectral information properly which has caused large spectral distortion. Based on the experiments, the proposed uSDN powered by the unique sparse Dirichlet-net outperforms all of the other approaches in terms of both RMSE and SAM, and it is quite stable for different types of input images. 

\begin{table}[htb]
	\centering
	\caption{Benchmarked results in terms of RMSE.}
	\label{tab:rmse}
	\begin{tabular}{p{1.0cm}| p{0.7cm} p{0.4cm} p{0.4cm} p{0.4cm} p{0.6cm}|p{0.4cm}p{0.6cm}}
		\hline
		\multirow{2}{*}{Methods}&\multicolumn{5}{|c|}{CAVE}&\multicolumn{2}{|c}{Harvard} \\
		\cline{2-8}
		{}&balloon&CD&cloth&photo&spool&img1&imgb5\\
		\hline
		CS &	25.4&	19.4&	22.0&	18.2&	25.8&16.7&17.8\\
		MRA&	12.5&	14.2&	15.4&	4.8&	11.3&4.7&8.9\\
		BS&	14.2&	15.3&	17.6&	11.3&	15.2&10.9&14.7\\
		Hysure&	14.9&	20.3&	14.8&	4.6&12.5&4.4&5.4\\
		BSR&	2.6 &7.9&	4.3&2.1&	6.2&2.3&2.5\\
		CNMF&	9.0&	11.9&	10.1&5.2&	12.2&3.2&4.5\\		
		CSU& 13.3&	10 &	6.7&3.1&7.9&2.2&2.6\\
		uSDN &\textbf{1.8}&\textbf{4.8}&\textbf{3.7}&\textbf{2.0}&\textbf{5.3}&\textbf{2.0}&\textbf{0.7}\\
		\hline
		HBPG& 1.9 &5.3&	3.7&--&	--&2.2&0.8\\		
		\hline
	\end{tabular}
\end{table}

\begin{table}
	\centering
	\caption{Benchmarked results in terms of SAM.}
	\label{tab:sam}
	\begin{tabular}{p{1.0cm}| p{0.7cm} p{0.4cm} p{0.4cm} p{0.4cm} p{0.6cm}|p{0.4cm}p{0.6cm}}
		\hline
		\multirow{2}{*}{Methods}&\multicolumn{5}{|c|}{CAVE}&\multicolumn{2}{|c}{Harvard} \\
		\cline{2-8}
		{}&balloon&CD&cloth&photo&spool&img1&imgb5\\
		\hline
		CS&	19&	17&	17&82&	48&15&14\\
		MRA&	12&	9&	11&	14&	15&13&15\\
		BS&	11&	16&	10&	18&	24&17&18\\
		Hysure&	18&	24&	18&	19&	38&18&19\\
		BSR&11.9&	17.9&	6&	14&	16&1.9&3.4\\
		CNMF&	10&	9&	7&	11&	20&10&13\\
		CSU&	8.9&	25&	12.6&	10&	17& 1.8&2.8\\
		uSDN&\textbf{4.7}&\textbf{10}&\textbf{4.8}&\textbf{5.4}&\textbf{13}&\textbf{1.6}&\textbf{1.7}\\
		\hline
		HBPG& 7.6 &10.6&5.0&--&	--&2.5&2.1\\		
		\hline
	\end{tabular}
\end{table}

\begin{table}
	\centering
	\caption{The average RMSE and SAM scores over complete benchmarked datasets.}
	\vspace*{-2mm}
	\label{tab:average}
	\begin{center}
		\begin{tabular}{l|cc|cc}
			\hline
			\multirow{2}{*}{Methods}&\multicolumn{2}{|c|}{CAVE}&\multicolumn{2}{|c}{Harvard} \\
			\cline{2-3}\cline{4-5}
			&RMSE&SAM&RMSE&SAM\\
			\hline
			CSU\cite{lanaras2015hyperspectral}&	9.96&15.63& 3.37&5.35\\		
			BSR\cite{akhtar2015bayesian}&	5.29 &13.63& 2.61&4.46\\		
			uSDN &\textbf{4.09}&\textbf{6.95}&\textbf{1.78}&\textbf{4.05}\\		 
			\hline
		\end{tabular}
	\end{center}
\vspace*{-5mm}
\end{table}
To further demonstrate the robustness of the proposed uSDN, we report the mean of RMSE and SAM over the complete CAVE and Harvard dataset in Table 3. We only list the performance of matrix factorization based CSU and Bayesian based BSR, since they demonstrated better performance as shown in Tables 1 and 2. We observe that since BSR estimates the representations separately from the spectral bases, although it can achieve good RMSE scores, its SAM scores are not promising.  While CSU relates the representations with a predefined down-sampling function, and thus achieves better results on the Harvard dataset, it generates worse results on the CAVE dataset. Both methods may cause spectral distortion in different scenarios. The proposed approach consistently outperforms the other methods in terms of both RMSE and SAM as reported in Table ~\ref{tab:average}. 

We also make two further observations. First, since the Harvard dataset is less challenging than the CAVE dataset, the improvement on the former is not as apparent as that on the latter. This, on the other hand, demonstrates that uSDN can handle challenging scenarios much better than state-of-the-art. Second, uSDN is very effective in preserving the spectral signature of the reconstructed HR HSI, showing much improved performance especially on SAM on CAVE. The main reason that contributes to the success of the proposed approach is that it relates the representations $\mathbf{S}_h$ and $\mathbf{S}_m$ with statistics and angular difference, \ie, both representations are encouraged to follow a Dirichlet distribution, and their angular difference is enforced to be small. In this way, both the reconstruction error and spectral distortion are effectively reduced. 
Since the representation is enforced to be sparse Dirichlet over each pixel, not the entire image, the proposed structure is capable of recovering different pixels individually. And the total number of the recovered samples, that equals the number of pixels, is large. This demonstrates the representation capacity of the proposed structure. 

\begin{figure}[t]
	\begin{center}
		\begin{minipage}{0.9\linewidth}
			{\includegraphics[width=0.22\linewidth]{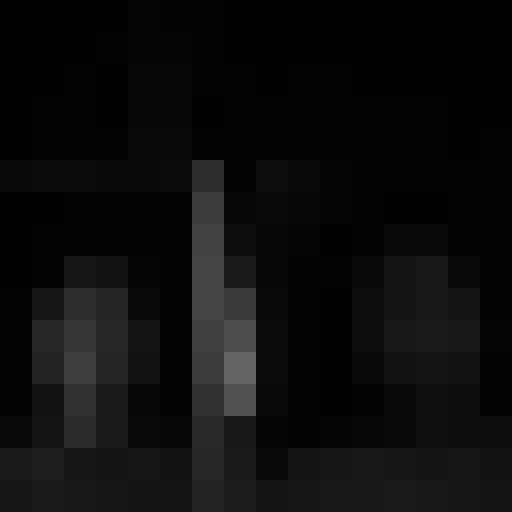}
				\label{fig:photo:a}}
			{\includegraphics[width=0.22\linewidth]{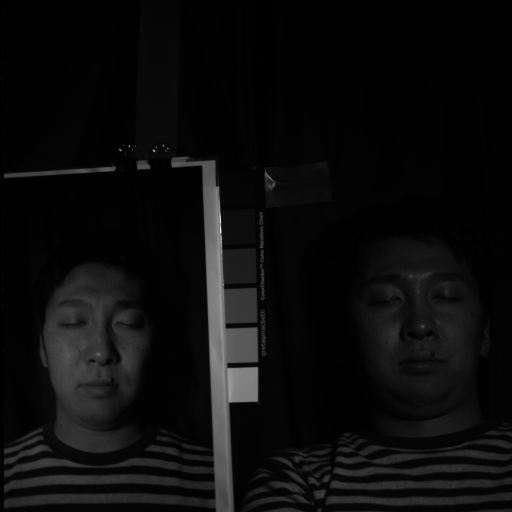}
				\label{fig:photo:b}}
			{\includegraphics[width=0.22\linewidth]{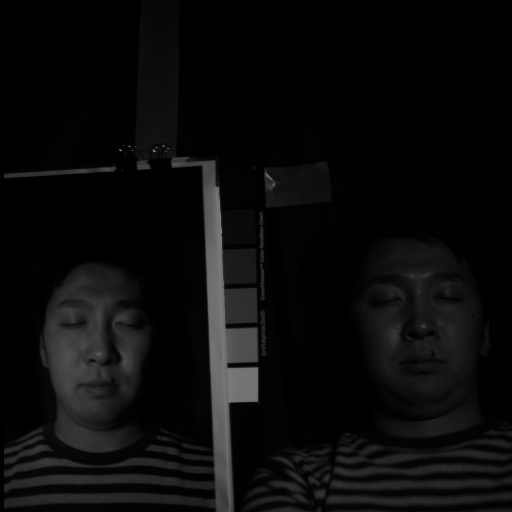}
				\label{fig:photo:c}}
			{\includegraphics[width=0.265\linewidth]{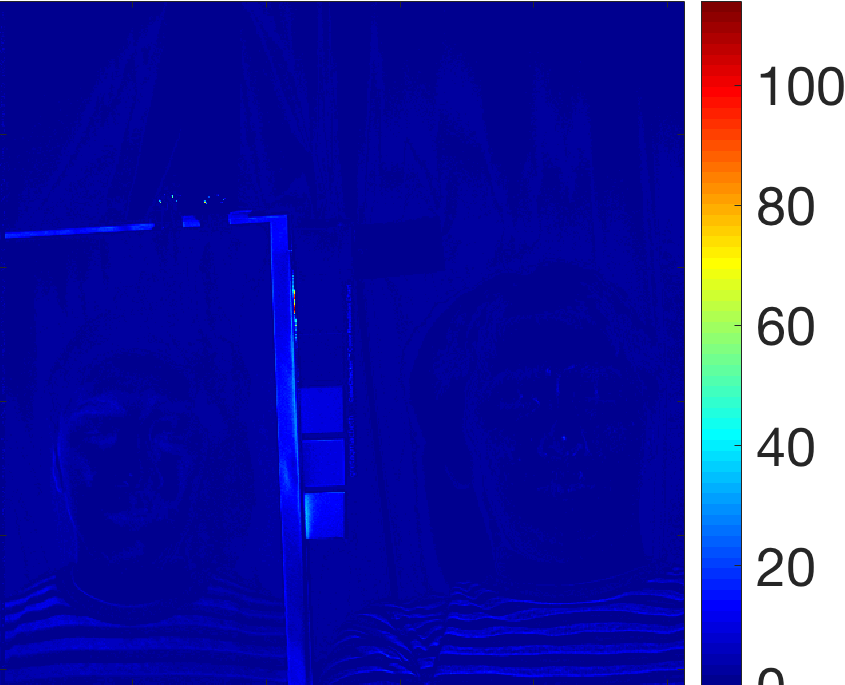}
				\label{fig:photo:d}}\\
			{\includegraphics[width=0.22\linewidth]{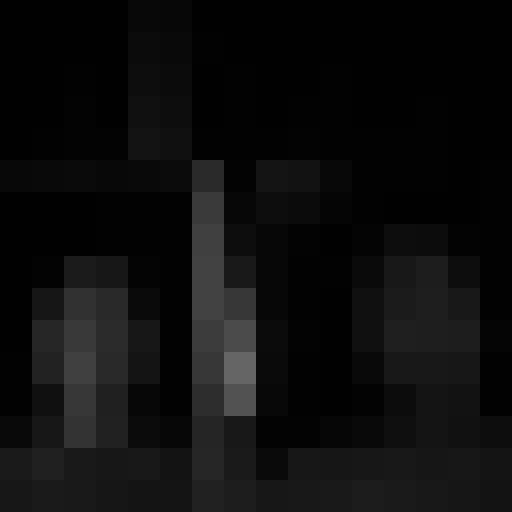}
				\label{fig:photo:e}}
			{\includegraphics[width=0.22\linewidth]{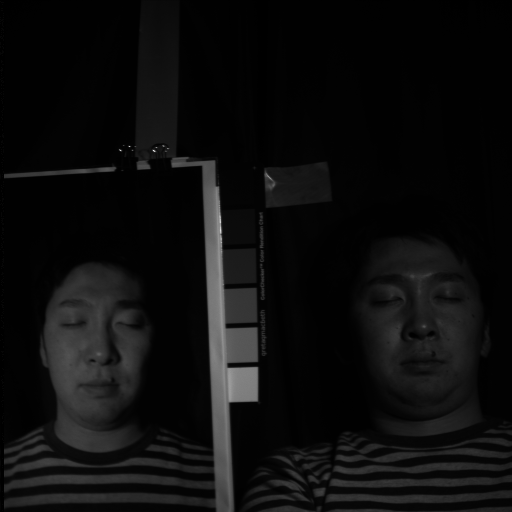}
				\label{fig:photo:f}}
			{\includegraphics[width=0.22\linewidth]{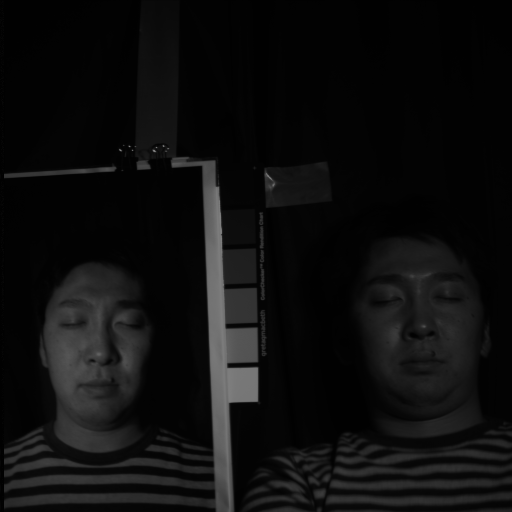}
				\label{fig:photo:g}}
			{\includegraphics[width=0.265\linewidth]{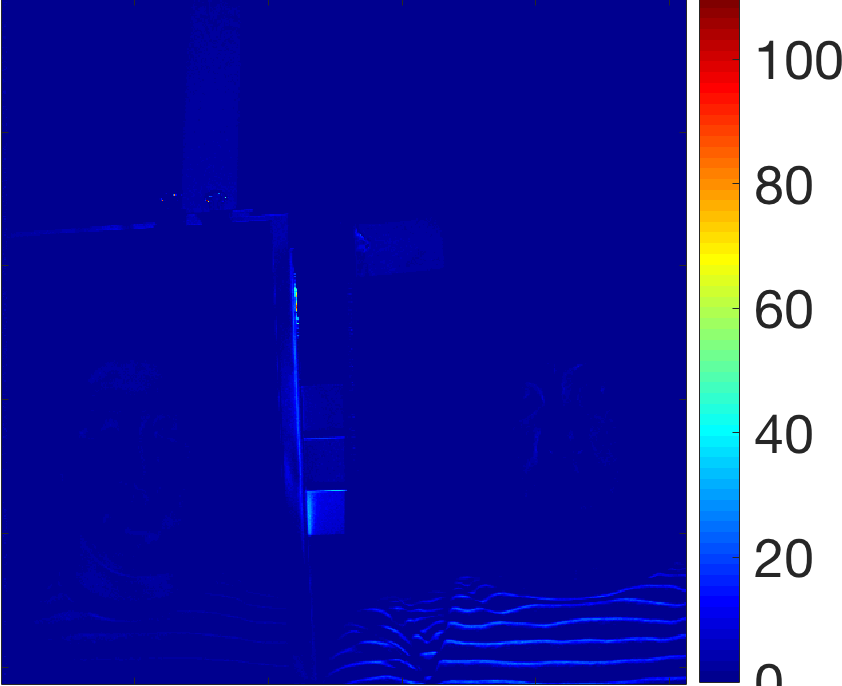}
				\label{fig:photo:h}}\\
			{\includegraphics[width=0.22\linewidth]{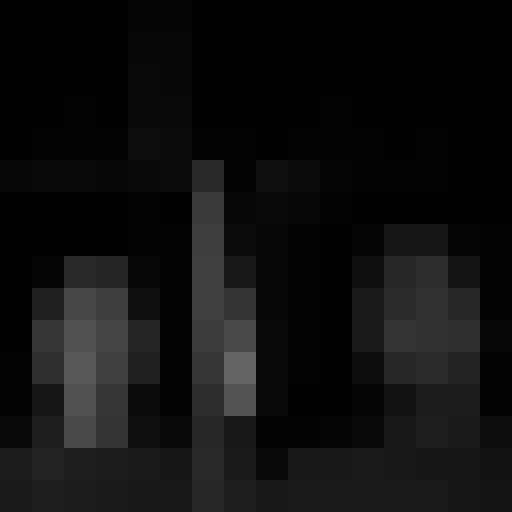}
				\label{fig:photo:i}}
			{\includegraphics[width=0.22\linewidth]{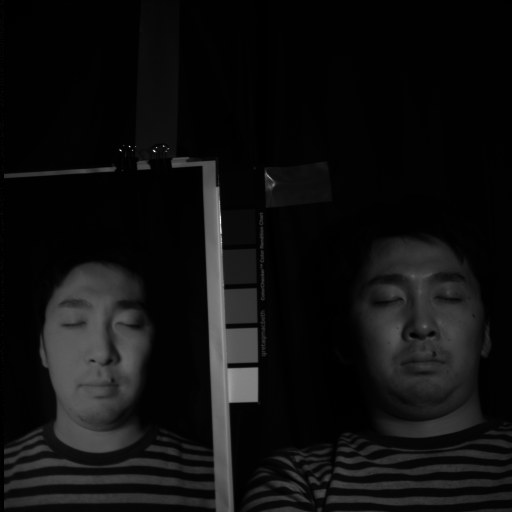}
				\label{fig:photo:j}}
			{\includegraphics[width=0.22\linewidth]{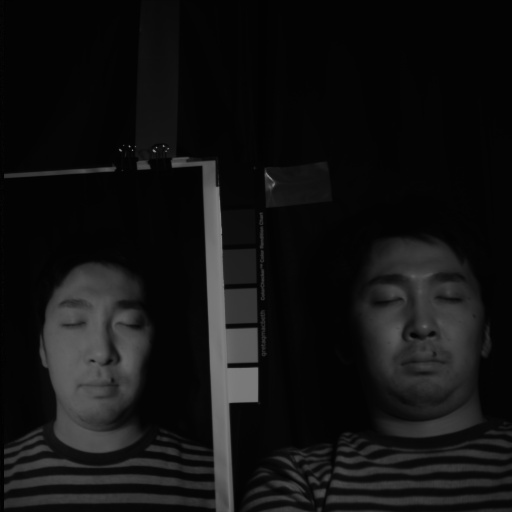}
				\label{fig:photo:k}}
			{\includegraphics[width=0.265\linewidth]{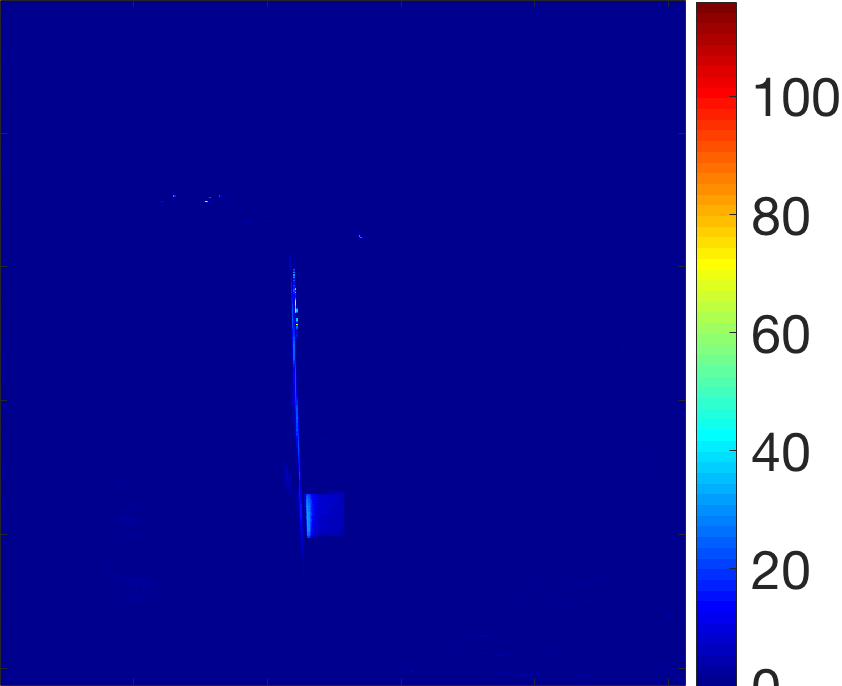}
				\label{fig:photo:l}}	
		\end{minipage}\vspace{3mm}
		\begin{minipage}{0.9\linewidth}
			{\includegraphics[width=0.22\linewidth]{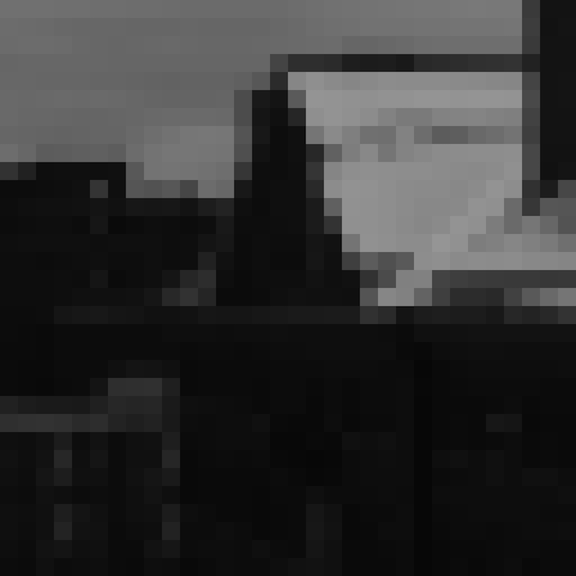}
				\label{fig:img1:a}}
			{\includegraphics[width=0.22\linewidth]{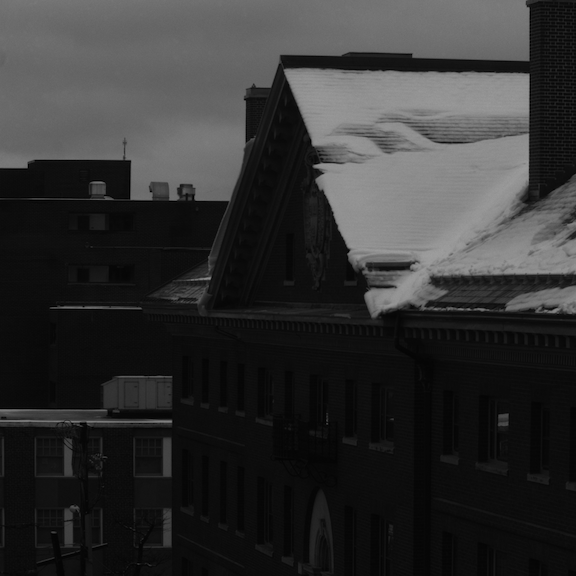}
				\label{fig:img1:b}}
			{\includegraphics[width=0.22\linewidth]{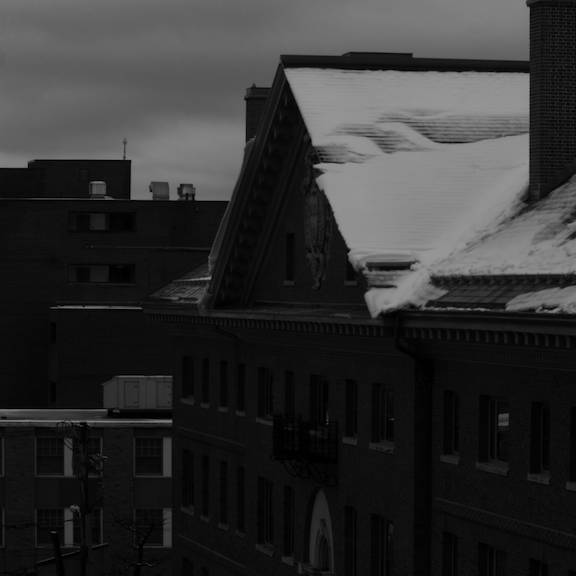}
				\label{fig:img1:c}}
			{\includegraphics[width=0.26\linewidth]{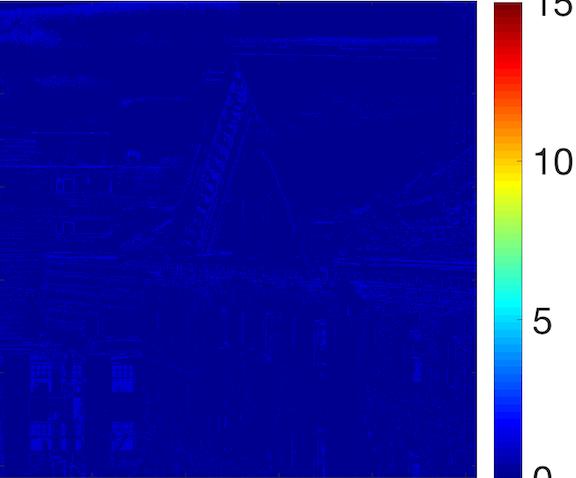}
				\label{fig:img1:d}}\\
			{\includegraphics[width=0.22\linewidth]{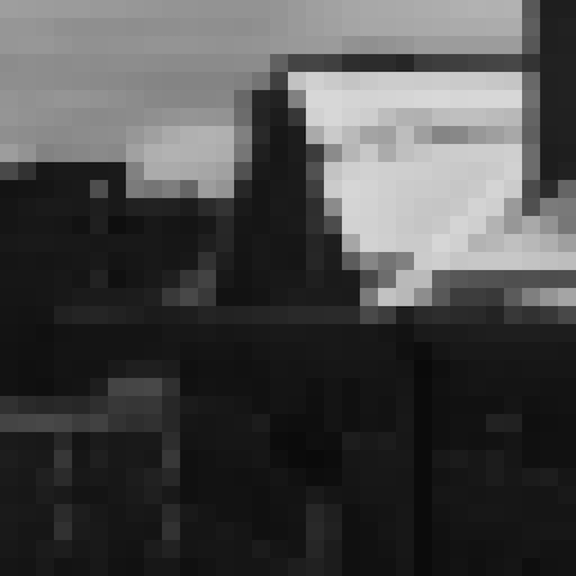}
				\label{fig:img1:e}}
			{\includegraphics[width=0.22\linewidth]{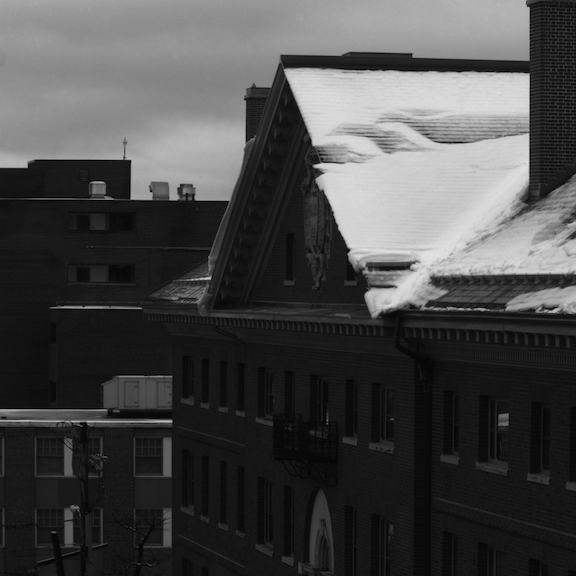}
				\label{fig:img1:f}}
			{\includegraphics[width=0.22\linewidth]{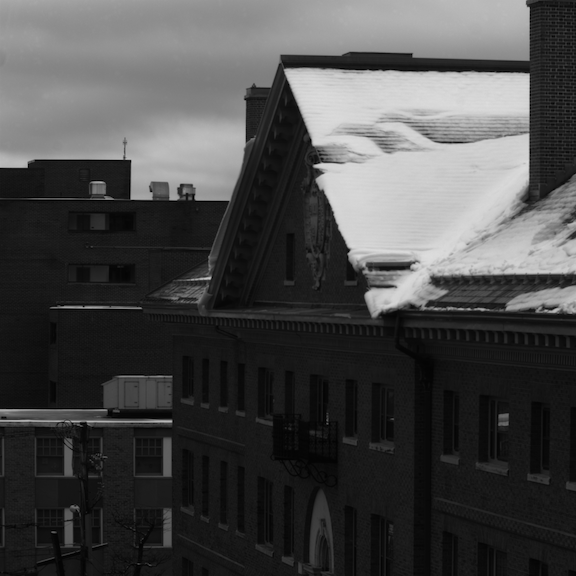}
				\label{fig:img1:g}}
			{\includegraphics[width=0.26\linewidth]{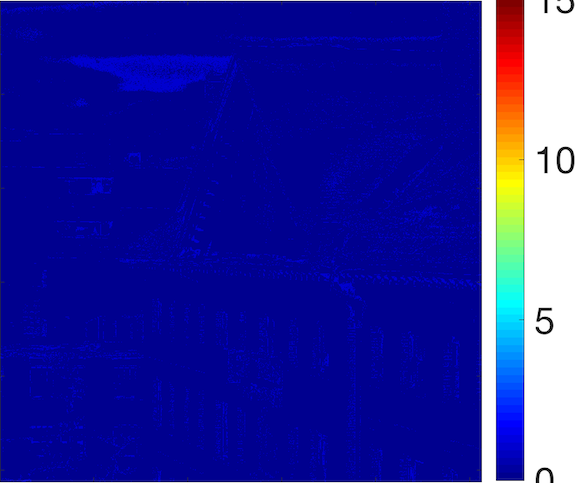}
				\label{fig:img1:h}}\\
			{\includegraphics[width=0.22\linewidth]{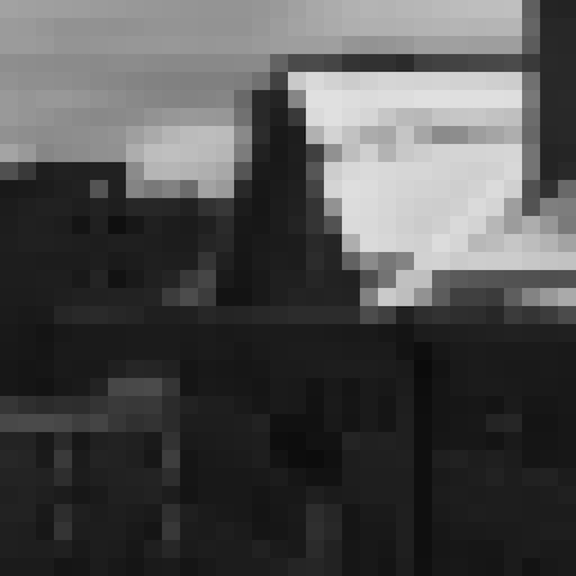}
				\label{fig:img1:i}}
			{\includegraphics[width=0.22\linewidth]{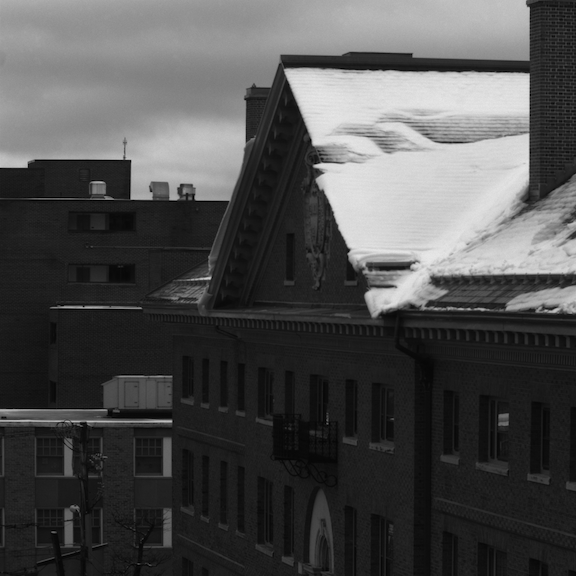}
				\label{fig:img1:j}}
			{\includegraphics[width=0.22\linewidth]{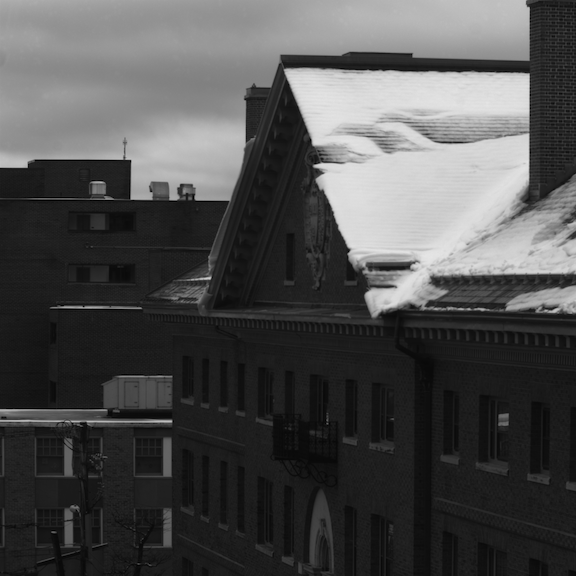}
				\label{fig:img1:k}}
			{\includegraphics[width=0.26\linewidth]{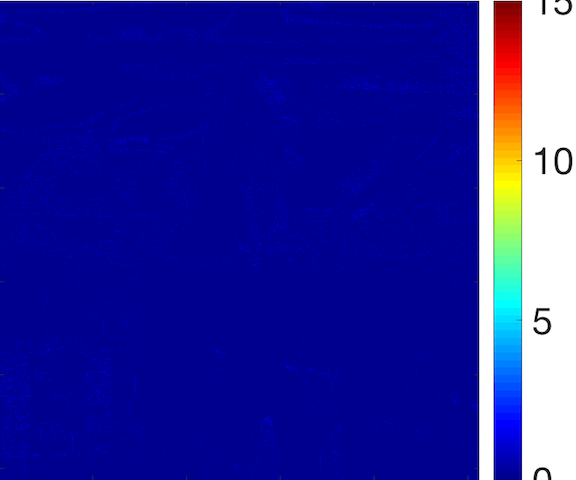}
				\label{fig:img1:l}}\\
		\end{minipage}
	\end{center}
	\vspace{-4mm}
	\caption{Reconstructed images from the CAVE  (top) and Harvard dataset (bottom) at wavelength 460, 540 and 620 nm. First column: LR images ($16\times16$). Second: estimated images ($512\times512$). Third: ground truth images. Fourth: absolute difference.}
	\label{fig:photo}
	\vspace{-4mm}
\end{figure}

\begin{figure}[t]
	\begin{center}
		\begin{minipage}{0.9\linewidth}
			\hspace{0mm}{\includegraphics[width=0.22\linewidth]{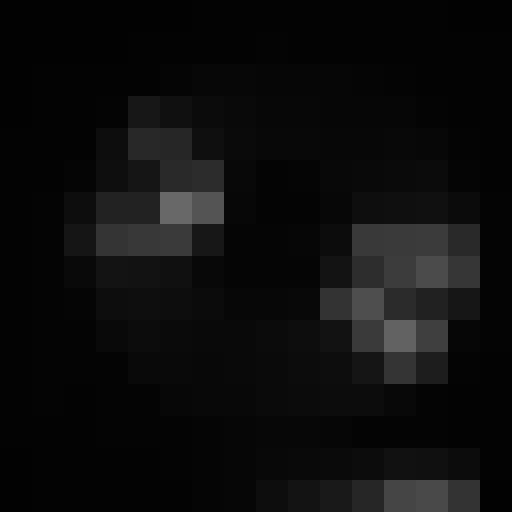}
				\label{fig:cd:a}}
			{\includegraphics[width=0.22\linewidth]{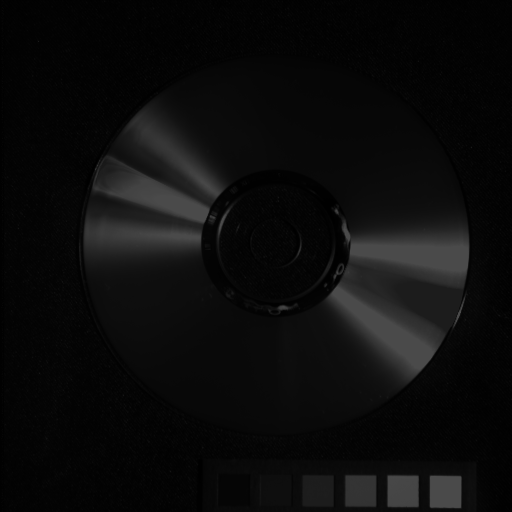}
				\label{fig:cd:b}}
			{\includegraphics[width=0.22\linewidth]{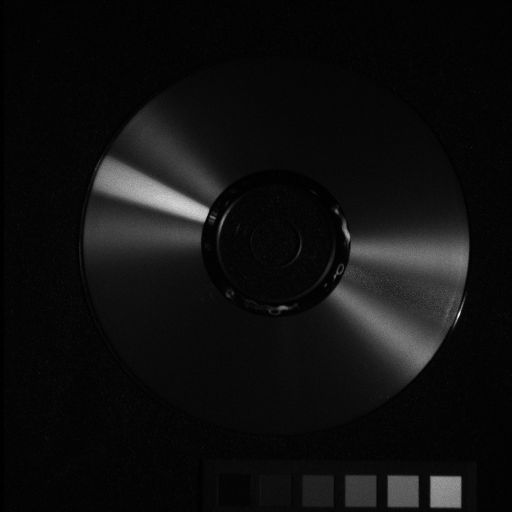}
				\label{fig:cd:c}}
			{\includegraphics[width=0.22\linewidth]{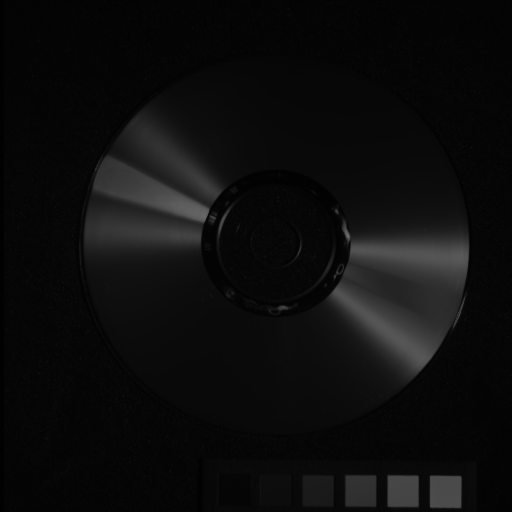}
				\label{fig:cd:d}}\\
			{\includegraphics[width=0.22\linewidth]{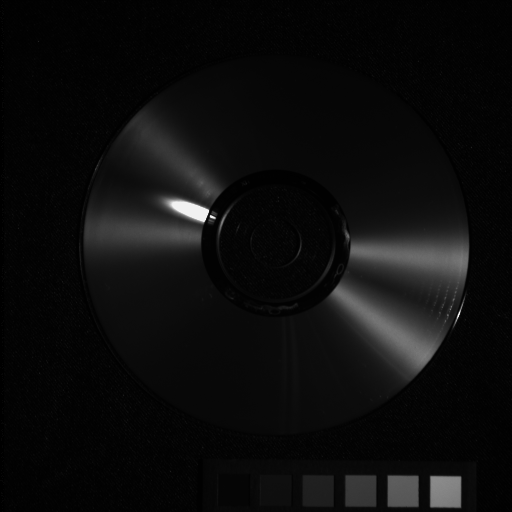}
				\label{fig:cd:e}}
			{\includegraphics[width=0.22\linewidth]{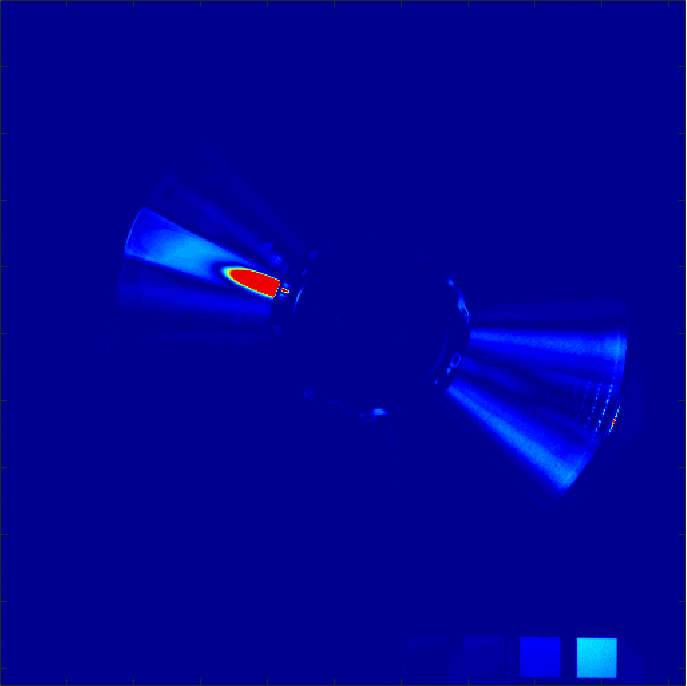}
				\label{fig:cd:f}}
			{\includegraphics[width=0.22\linewidth]{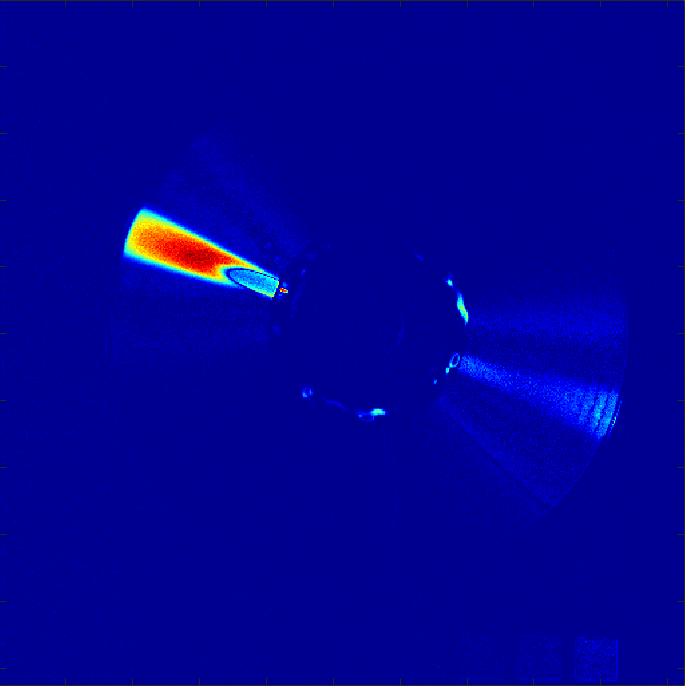}
				\label{fig:cd:g}}
			{\includegraphics[width=0.27\linewidth]{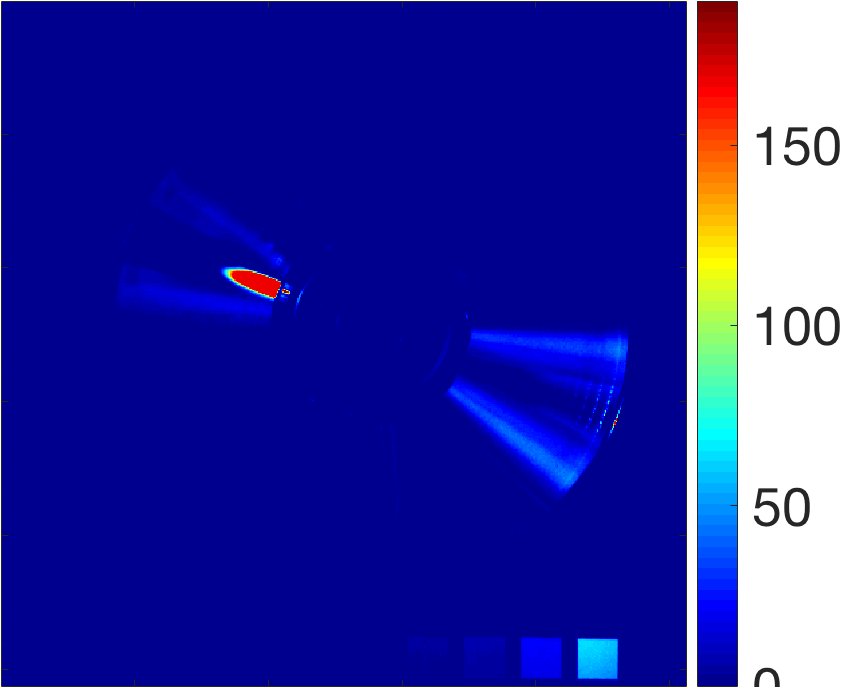}
				\label{fig:cd:h}}
		\end{minipage}
		\begin{minipage}{0.9\linewidth}
			\hspace{0mm}{\includegraphics[width=0.22\linewidth]{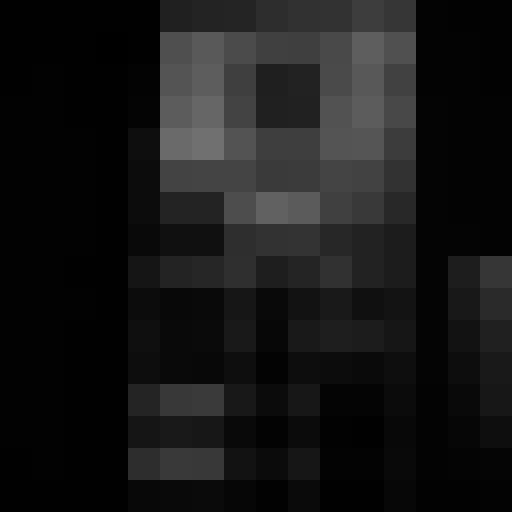}
				\label{fig:spool:a}}
			{\includegraphics[width=0.22\linewidth]{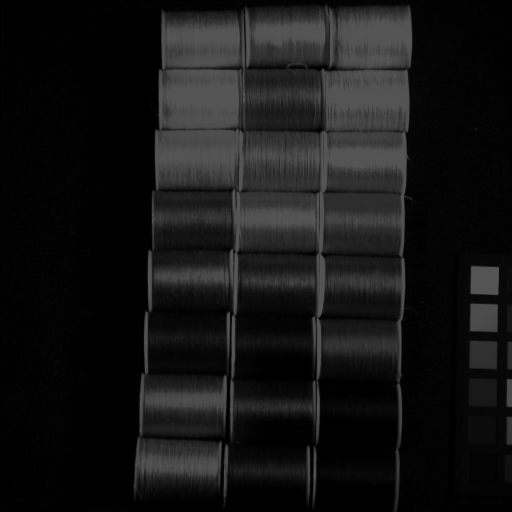}
				\label{fig:spool:b}}
			{\includegraphics[width=0.22\linewidth]{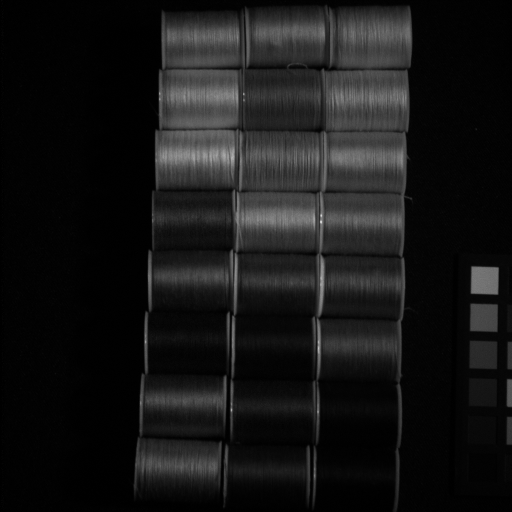}
				\label{fig:spool:c}}
			{\includegraphics[width=0.22\linewidth]{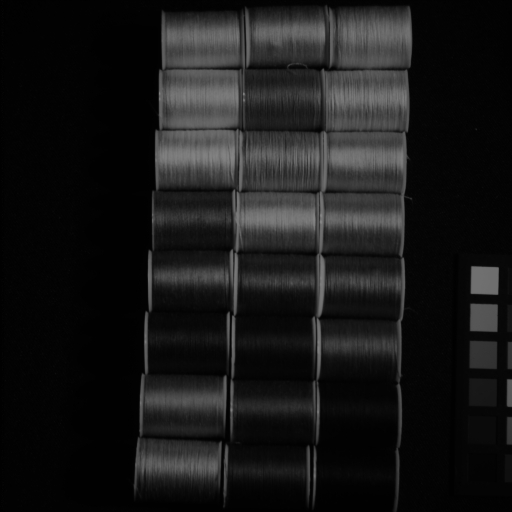}
				\label{fig:spool:d}}\\
			{\includegraphics[width=0.22\linewidth]{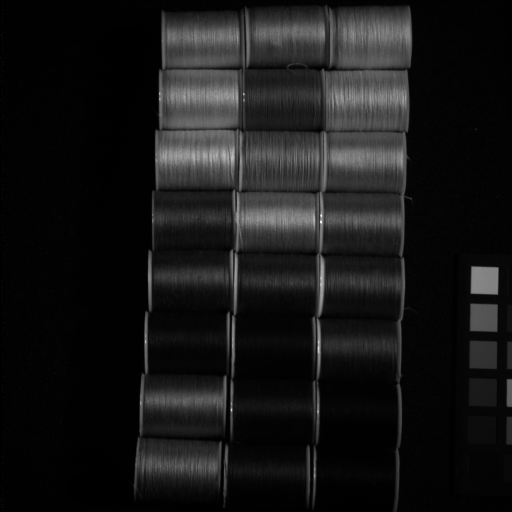}
				\label{fig:spool:e}}
			{\includegraphics[width=0.22\linewidth]{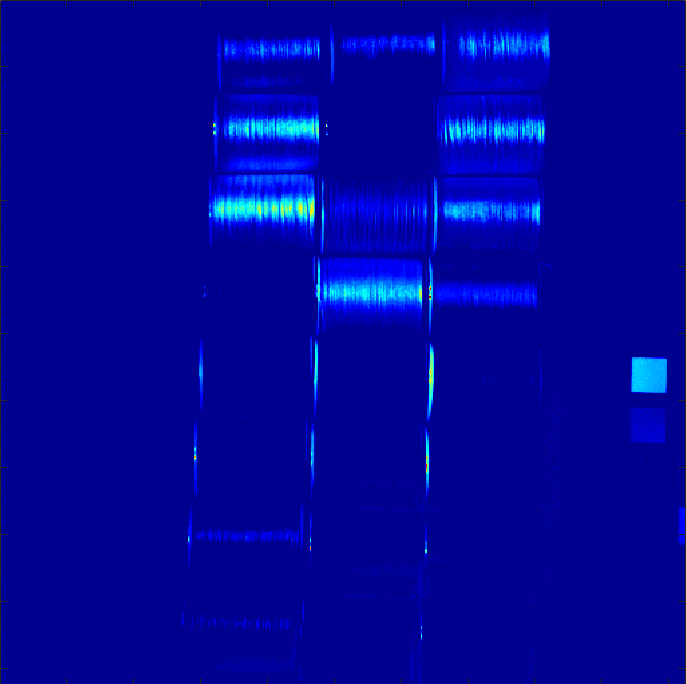}
				\label{fig:spool:f}}
			{\includegraphics[width=0.22\linewidth]{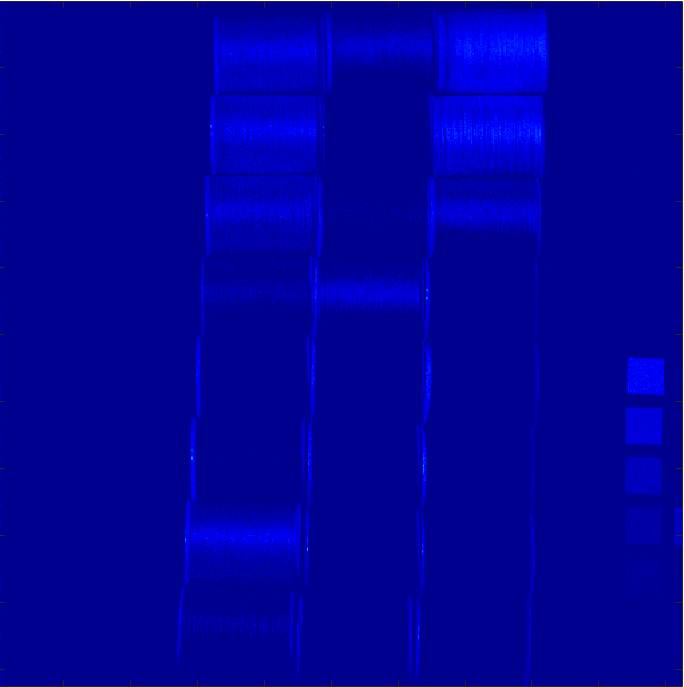}
				\label{fig:spool:g}}
			{\includegraphics[width=0.27\linewidth]{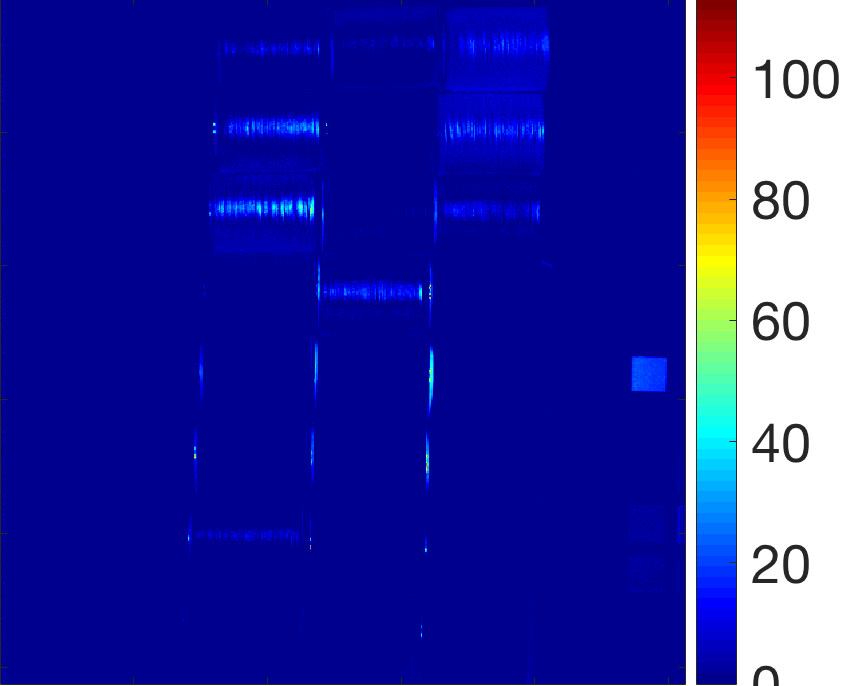}
				\label{fig:spool:h}}	
		\end{minipage}
	\end{center}
	\vspace{-4mm}
	\caption{Reconstructed images of two examples (top two rows and bottom two rows) from the CAVE dataset at wavelength 670 nm. The first column shows the LR image (top) and the ground truth image (bottom). The second, third and fourth columns are the reconstructed results (top) and the absolute difference (bottom) from CSU, BSR and uSDN, respectively.}
	\label{fig:cd}
\vspace{-4mm}
\end{figure}

To visualize the results, we show the reconstructed samples from CAVE and Harvard taken at wavelengths 460, 540, and 670 nm in Fig.~\ref{fig:photo}. The first through fourth columns show the LR images, reconstructed images from our method, ground truth images, and the absolute difference between the images at the second and third columns, respectively. We also compare the proposed method with CSU and BSR on the challenge dataset CAVE and show the results in Fig.~\ref{fig:cd}. The effectiveness of the proposed method can be readily observed from the difference images, where the proposed approach is able to preserve both the spectral and spatial information. 

\begin{figure}[t]
	\begin{center}
		\begin{minipage}{0.45\linewidth}
			{\includegraphics[width=0.9\linewidth]{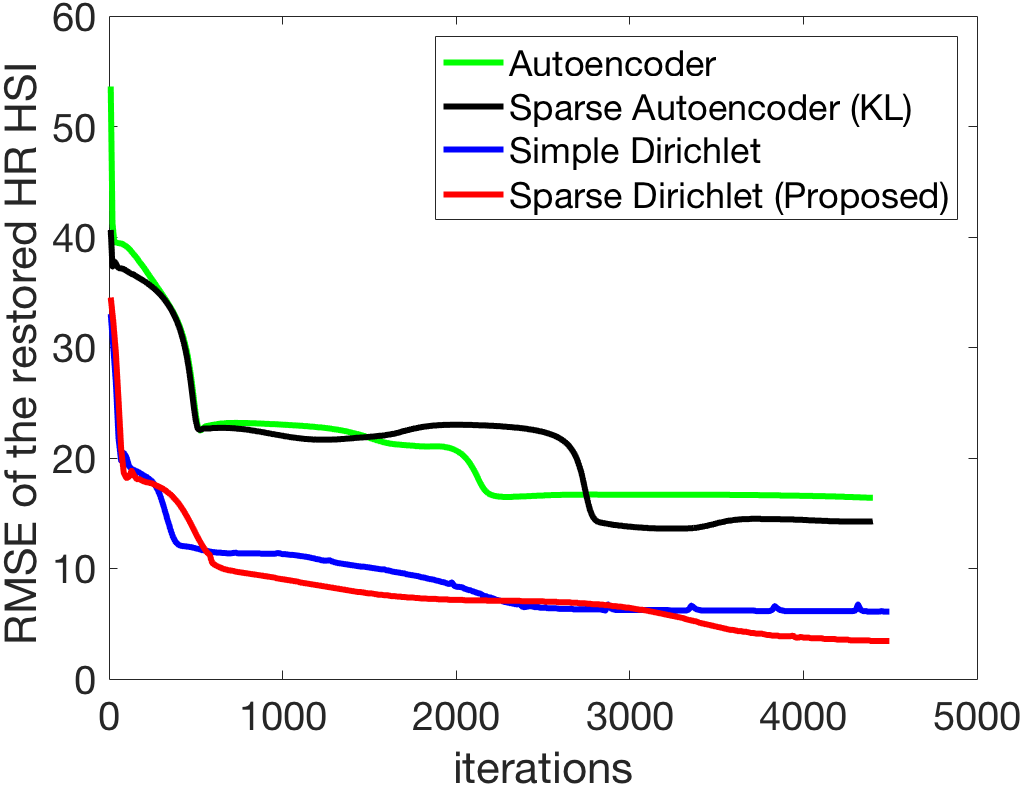}\label{fig:dir}}
			\caption{Sparse Dirichlet constraint.}
			\label{fig:sparse}
		\end{minipage}\hfill	
		\begin{minipage}{0.45\linewidth}
			{\includegraphics[width=0.9\linewidth]{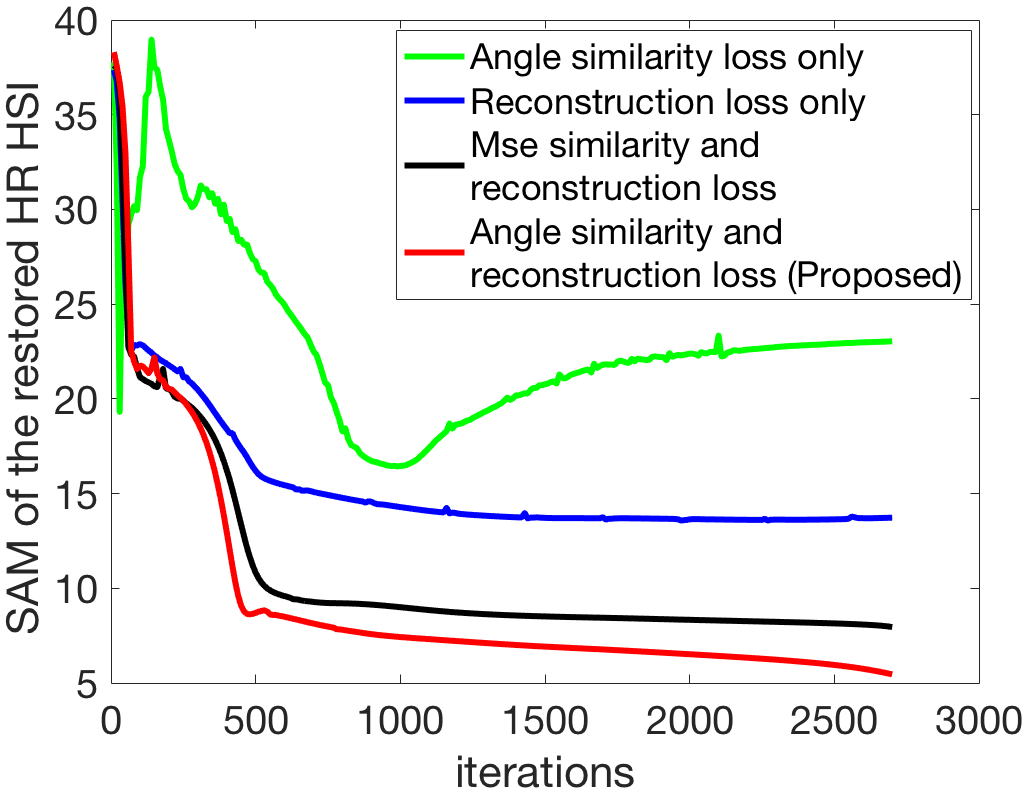}\label{fig:angle}}\hfill
			\caption{Spectral angle constraint.}
			\label{fig:spectral}
		\end{minipage}\hfill\\
		\begin{minipage}{0.3\linewidth}
			{\includegraphics[width=0.9\linewidth]{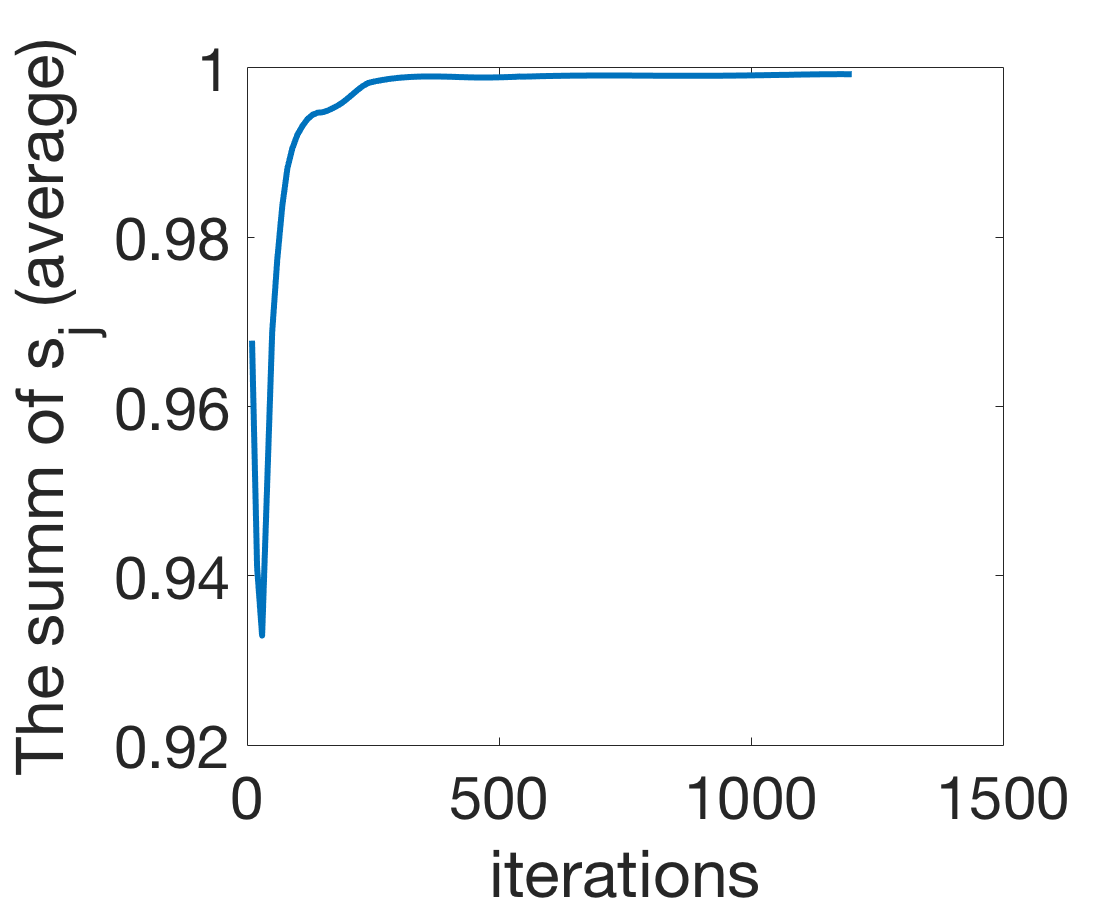}}
			\caption{Summation.}
			\label{fig:sum}
		\end{minipage}\hfill
		\begin{minipage}{0.33\linewidth}
			{\includegraphics[width=0.9\linewidth]{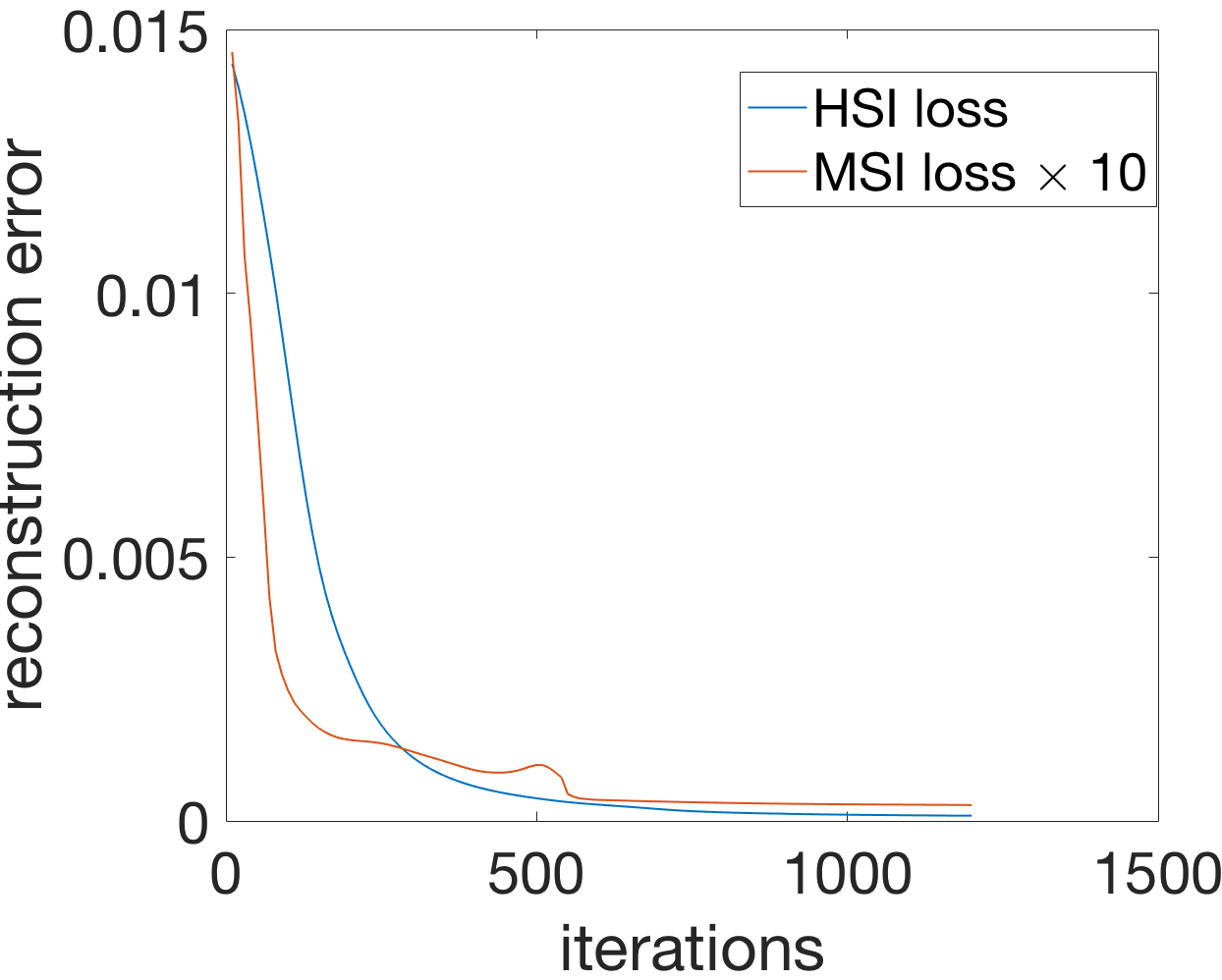}}
			\caption{Learning curves.}
			\label{fig:loss}
		\end{minipage}\hfill
		\begin{minipage}{0.33\linewidth}
			{\includegraphics[width=0.9\linewidth]{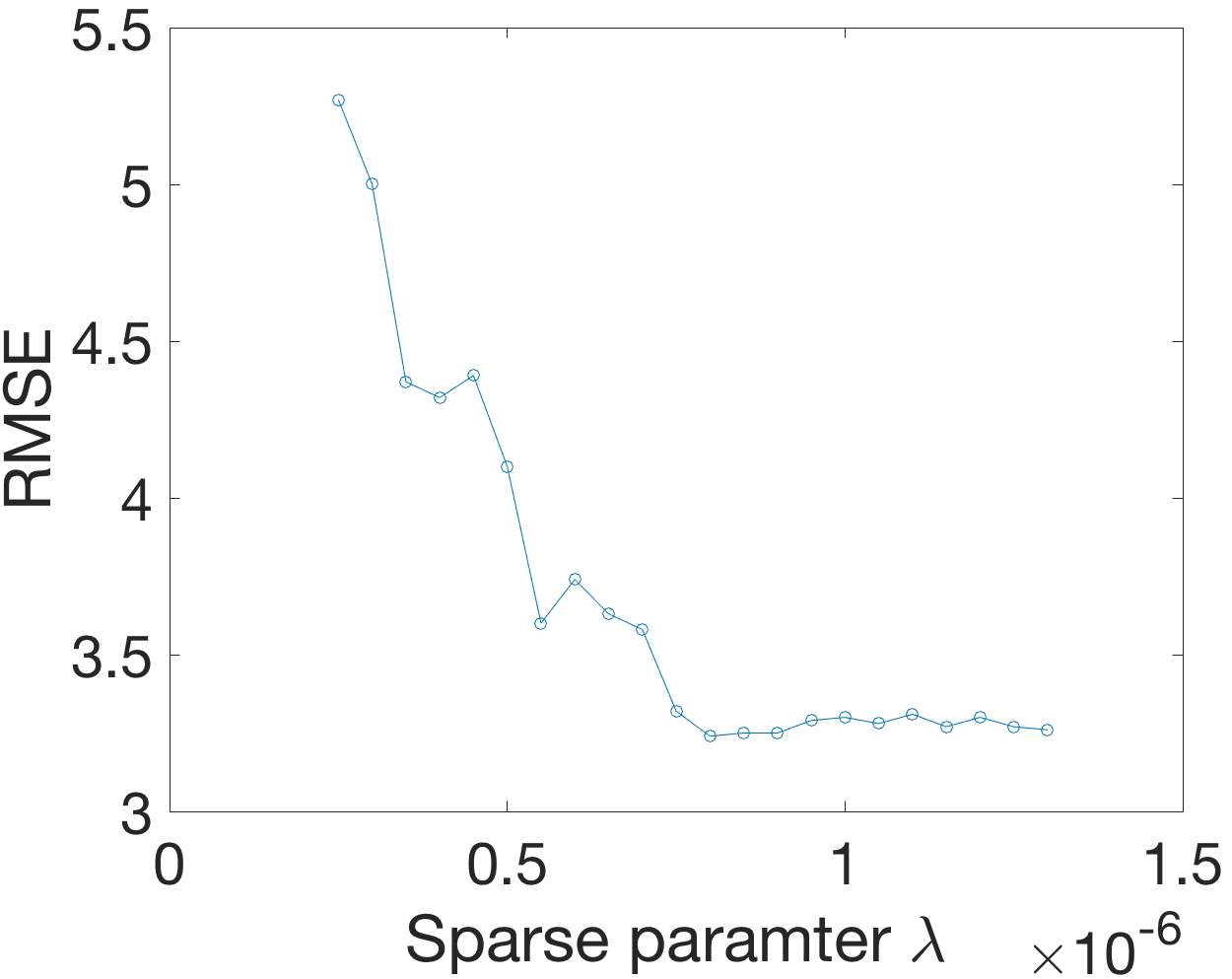}}
			\caption{The RMSE curve.}
			\label{fig:para}
		\end{minipage}\hfill\\		
		\begin{minipage}{0.45\linewidth}
			{\includegraphics[width=0.9\linewidth]{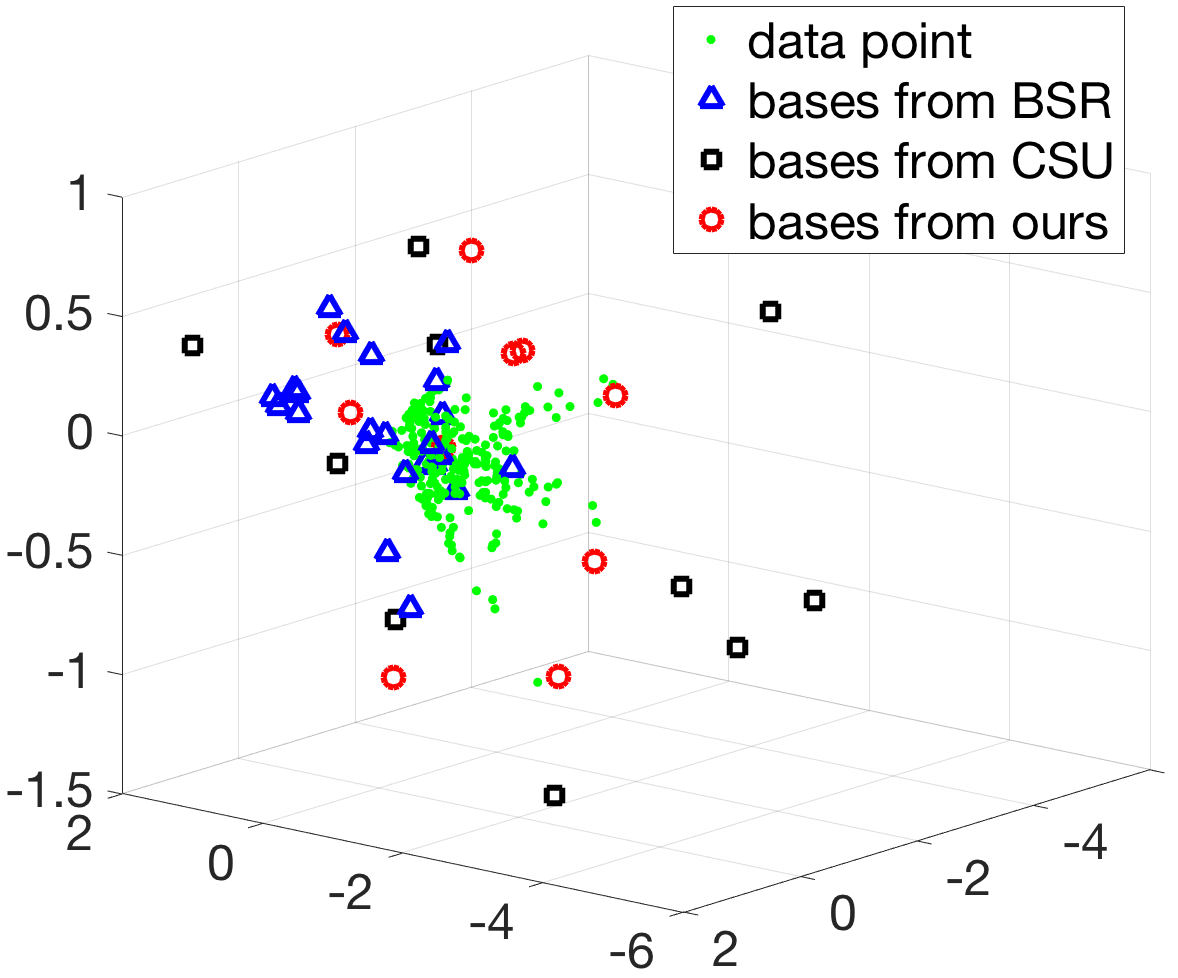}}
			\caption{Spectral basis.}
			\label{fig:basis}
		\end{minipage}\hfill		
		\begin{minipage}{0.45\linewidth}
			{\includegraphics[width=0.9\linewidth]{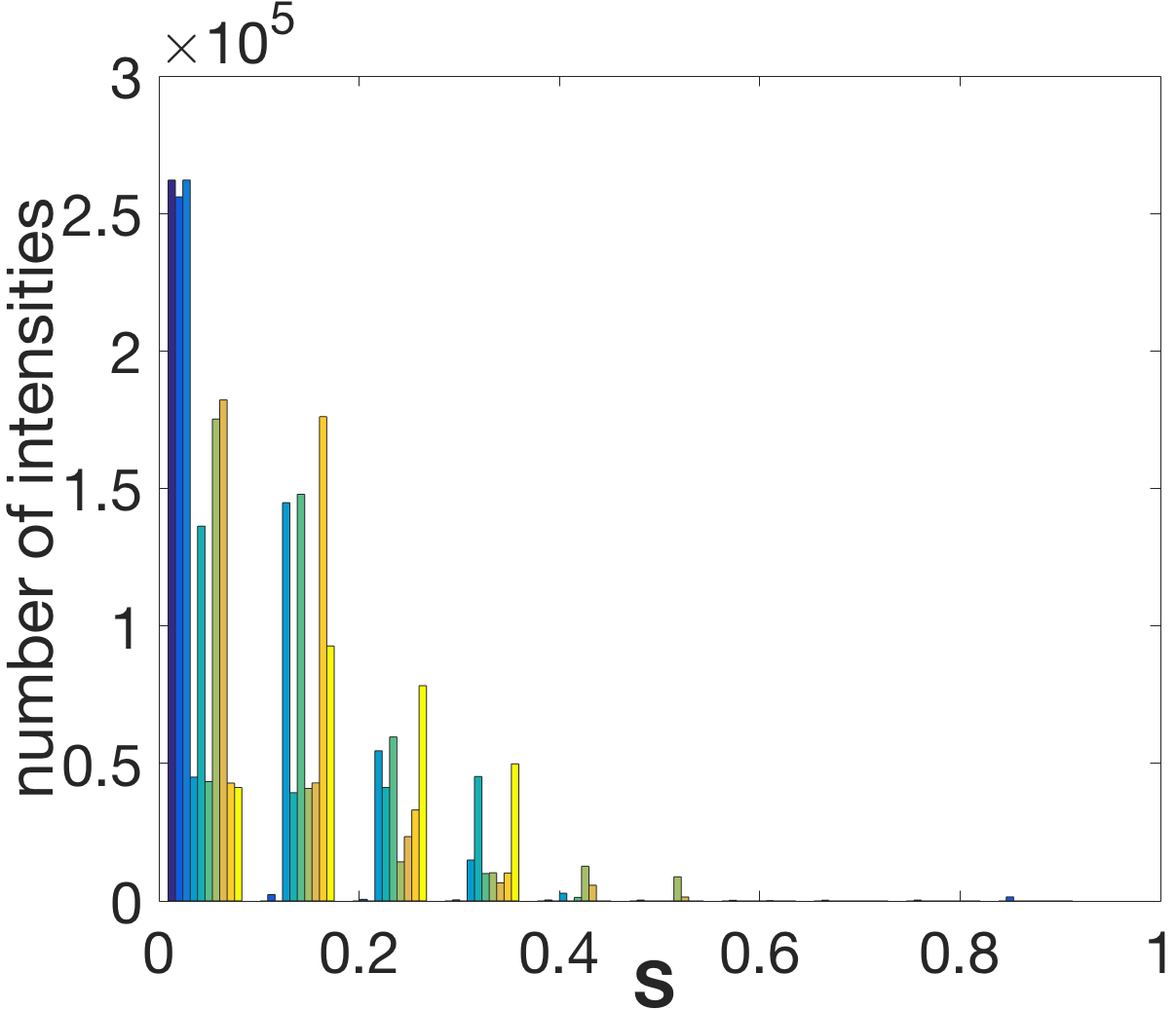}}
			\caption{Histogram of $\mathbf{S}_m$.}
			\label{fig:hist}
		\end{minipage}\hfill
	\end{center}
\vspace{-5mm}
\end{figure}

\textbf{Ablation Study:} Taking the 'pompom' image from the CAVE dataset as an example, we further evaluate 1) the necessity of enforcing the representation to follow sparse Dirichlet and 2) the usage of angle similarity loss. Fig. \ref{fig:sparse} illustrates the RMSE of the reconstructed HR HSI using 4 different network structures, \ie, autoencoder (AE) without any constraints, AE with the sparsity constraint (SAE), a simple Dirichlet-Net without any constraints, and the proposed sparse Dirichlet-Net. We observe that the adoption of Dirichlet-Net significantly reduces RMSE as compared to AE and SAE; and the proposed sparse Dirichlet-Net reduces RMSE even further, especially as the number of iterations increases. Fig. \ref{fig:sum} shows the summation of elements in $\mathbf{s}_j$ averaged over all pixels in the image, where we observe that representations $\mathbf{s}_j$ are sum-to-one almost surely after around 300 iterations with Dirichlet-Net. Fig. \ref{fig:spectral} demonstrates the spectral angle mapper (SAM) of the reconstructed HR HSI using 4 different loss functions when the MSI network is updated, \ie, only with angle similarity loss, only with reconstruction loss, reconstruction loss with MSE similarity, and the proposed reconstruction loss with angle similarity, respectively. We observe that reconstruction loss significantly stabilizes/regulates the convergence process; and reconstruction loss with angle similarity presents the lowest SAM and fastest convergence speed.

\textbf{Convergence Study:} During the optimization, both the HSI and MSI networks converge smoothly as shown in Fig.~\ref{fig:loss}. The MSI network has a little bit fluctuation caused by the angular difference which is minimized every 10 iterations between the representations of two modalities. 

\textbf{Effect of Free Parameters:} 
There are two free parameters in the algorithm design, i.e., $\mu$ for the decoder weight loss and $\lambda$ for the sparsity control, as shown in Eq.~\eqref{equ:objhsi}. We keep $\mu=1e^{-6}$ during the experiments. Fig.~\ref{fig:para} shows how RMSE is decreasing when we increase $\lambda$ from $2\times10^{-7}$ to $1\times10^{-6}$. We set $\lambda=1\times10^{-6}$ in the experiments. 

\textbf{Visualizing $\mathbf{S}_m$ and $\mathbf{\Phi}_h$:} The proposed structure is based on the assumption that the LR HSI, HR MSI, and HR HSI can be formulated as a linear combination of their corresponding spectral bases. Here, we would like to provide visualization results of the spatial representation, $\mathbf{S}_m$, its sparsity property, and the spectral bases, $\mathbf{\Phi}_h$. We use the pompom image from the CAVE dataset as the testing image to generate all the visualization. In order to visually see if the linear combination assumption is valid or not, we project the estimated bases, $\mathbf{\Phi}_m$ into a 3D space using singular value decomposition. In Fig.~\ref{fig:basis}, we observe that the learned bases from CSU is a little bit far away from the data, while the bases from BSR cluster with each other and do not cover all the data. The bases from our method circumscribe the entire data, indicating a more effective representation of the data. We also study if $\mathbf{S}_m$ is indeed sparse or not. The histogram of the learned representations $\mathbf{S}_m$ is shown in Fig.~\ref{fig:hist}, where the sparsity is clearly evident. 


\section{Conclusion}
We proposed an unsupervised sparse Dirichelet-Net (uSDN) to solve the problem of hyperspectral image super-resolution (HSI-SR). To the best of our knowledge, this is the first effort to solving the problem of HSI-SR in an unsupervised fashion. The network extracts the spectral basis from LR HSI with rich spectral information and spatial representations from HR MSI with high spatial information through a shared decoder. The representations from two modalities are encouraged to follow a sparse Dirichlet distribution. In addition, the angular difference of two representations is minimized during the optimization to reduce spectral distortion. Extensive experiments on two benchmark datasets demonstrate the superiority of the proposed approach over state-of-the-art. 

\textbf{Acknowledgement:} This work was supported in part by NASA NNX12CB05C and NNX16CP38P.

{\small
	\bibliographystyle{ieee}
	\bibliography{cvpr_final}
}

\end{document}